\documentclass[review]{elsarticle}

\makeatletter
\def\ps@pprintTitle{%
 \let\@oddhead\@empty
 \let\@evenhead\@empty
 \def\@oddfoot{}%
 \let\@evenfoot\@oddfoot}
\makeatother

\usepackage{lineno,hyperref}
\usepackage{mathptmx,helvet,courier,makeidx,multicol,footmisc}
\usepackage{amsfonts,ifthen,color}
\usepackage{natbib}
\usepackage{times,setspace}

\usepackage{amssymb,latexsym,amsmath,amsthm}
\usepackage{graphicx}
\usepackage[table, xcdraw]{xcolor}
\usepackage{lscape}
\usepackage{algorithm, algpseudocode}
\usepackage{url}
\usepackage{mathtools}

\usepackage{tikz} % drawing graph
\usetikzlibrary{arrows.meta} % drawing graph
\usetikzlibrary{decorations.markings}
\usetikzlibrary{backgrounds}%https://www.overleaf.com/9631425144hzrbstvfqxkd
\usetikzlibrary{positioning,chains,fit,shapes,calc}
\usepackage{subcaption}
\usepackage{pdflscape,array,booktabs}

\renewcommand{\cite}{\citep}

\newcommand{\be}{\begin{equation}}
\newcommand{\ee}{\end{equation}}
\newcommand{\bd}{\begin{description}}
\newcommand{\ed}{\end{description}}
\newcommand{\ben}{\begin{enumerate}}
\newcommand{\een}{\end{enumerate}}
\newcommand{\beq}{\begin{quote}}
\newcommand{\eeq}{\end{quote}}
\newcommand{\bi}{\begin{itemize}}
\newcommand{\ei}{\end{itemize}}
\newcommand{\bea}{\begin{eqnarray}}
\newcommand{\eea}{\end{eqnarray}}
\newcommand{\bua}{\begin{eqnarray*}}
\newcommand{\eua}{\end{eqnarray*}}

\newcommand{\ba}{\begin{array}}
\newcommand{\ea}{\end{array}}
\newcommand{\bfig}{\begin{figure}}
\newcommand{\efig}{\end{figure}}
\newcommand{\bc}{\begin{center}}
\newcommand{\ec}{\end{center}}
\newcommand{\bt}{\begin{table}}
\newcommand{\et}{\end{table}}
\newcommand{\btab}{\begin{tabular}}
\newcommand{\etab}{\end{tabular}}

\renewcommand{\vec}[1]{\hat{\mathbf{#1}}}  % normal \vec not working

\DeclareMathOperator*{\argmin}{arg\,min}
\DeclarePairedDelimiter\floor{\lfloor}{\rfloor}

\newcommand{\indep}{~\raisebox{-0.3ex}{\rotatebox{90}{\ensuremath{\models}}}}
\newcommand{\dep}{\not \hspace*{-1.2ex}\indep}
\newcommand{\newspecial}[1]{
        \ifnum\includefigs>0
                \special{#1}
        \fi
}

\newboolean{draftp}

\newcommand{\newcomment}[1]{
\ifthenelse{\boolean{draftp}}{#1}{}
}

\definecolor{cyan}{rgb}{0.0, 1.0, 1.0}
\definecolor{babypink}{rgb}{0.96, 0.76, 0.76}

\newtheorem{proposition}{Proposition}[section]

\newtheorem{definition}{Definition}[section] % definition numbers are dependent on theorem numbers

\biboptions{authoryear}

\begin{document}

\begin{frontmatter}

\title{Markov Blanket Discovery using Minimum Message Length}

\author{Yang Li\corref{cor1}%
\fnref{fn1}}
\ead{kelvin.li@anu.edu.au}
\address{School of Computing, The Australian National University, Canberra, ACT, 2600, Australia}
\author{Kevin B Korb}
%\ead{cvr@sayahna.org}
\address{Faculty of Information Technology, Monash University, Clayton, VIC, 3800, Australia}
\author{Lloyd Allison}
%\ead[url]{www.stmdocs.in}
\address{Faculty of Information Technology, Monash University, Clayton, VIC, 3800, Australia}

\cortext[cor1]{Corresponding author}
%\fntext[fn1]{This is the first author footnote.}
%\fntext[fn2]{Another author footnote, this is a very long
%footnote and it should be a really long footnote. But this
%footnote is not yet sufficiently long enough to make two lines
%of footnote text.}

%\author{Yang Li\fnref{myfootnote}, Kevin B Korb, Lloyd Allison}
%\address{Faculty of Information Technology, Monash University, Clayton, Australia}


%\cortext[mycorrespondingauthor]{Yang Li (Kelvin)}
%\ead{kelvin.li@anu.edu.au}

\begin{abstract}
Causal discovery automates the learning of causal Bayesian networks from data and has been of active interest from their beginning. With the sourcing of large data sets off the internet, interest in scaling up to very large data sets has grown. One approach to this is to parallelize search using Markov Blanket (MB) discovery as a first step, followed by a process of combining MBs in a global causal model. We develop and explore three new methods of MB discovery using Minimum Message Length (MML) and compare them empirically to the best existing methods, whether developed specifically as MB discovery or as feature selection.  Our best MML method is consistently competitive and has some advantageous features.
\end{abstract}

%------------------------------------------------------------------------------------------------------
%------------------------------------------------------------------------------------------------------


\end{frontmatter}

%\linenumbers

%------------------------------------------------------------------------------------------------------
%------------------------------------------------------------------------------------------------------

\section{Introduction}
\label{sec:intro}

Causal discovery aims to learn causal Bayesian networks by using
information about conditional dependencies between sets of variables
gleaned from sample data. The research behind it is coterminous with
research on Bayesian networks themselves, beginning with
\citet{glymour1987}, spurred on by the difficulty and cost of
eliciting Bayesian networks from experts. The techniques behind causal
discovery have since then become more varied and effective, expanding
from the ``constraint-based'' learning of early efforts, which examine
conditional dependencies in isolation, to ``metric-based'' learning,
which apply Bayesian (or similar) metrics to causal models and data
sets as a whole. Meanwhile, however, the challenge has itself
increased manyfold, in particular, through the additional challenge of
``big data'' driven by the expansion of the internet.

This challenge is also aggravated by the fact that causal discovery itself
is NP-hard \citep{chickering1994learning}, forcing the use of
heuristics when learning whole causal models. An alternative to
learning causal models globally is the Local-to-Global (LGL) paradigm
that first finds local structures within subsets of variables then
unifies them into a global structure.

A promising approach to an LGL technique is to parallelize search
using Markov Blanket (MB) discovery as the first step, discovering MBs
centered around each variable, and then gluing them all together.  Here
we support this approach, developing MB learning methods using Minimum
Message Length (MML) scores and comparing them experimentally with the
strongest alternative MB learning algorithms. While we did this with
the ultimate goal of learning global causal models, in this paper we
limit ourselves to the first step only, MB discovery. 

In Section \ref{sec:related} we first quickly review the main alternatives that have been
proposed and tested on MB learning. After developing our terminology in Section \ref{sec:mb},
we then introduce the relevant concepts for MML learning in Section \ref{sec:mml} and follow by
introducing three models for representing Markov Blankets using MML
and algorithms for learning them in Section \ref{sec:mbmml}. We finish by experimentally
comparing these with each other and with the alternatives, including a
discussion of their time complexity in Section \ref{sec:experiments}.

%--------------------------------------------------------------------------------------------------------------
%--------------------------------------------------------------------------------------------------------------

%------------------------------------------------------------------------------------------------------
%------------------------------------------------------------------------------------------------------

\section{Related work} 
\label{sec:related}

\begin{figure}[t]
\centering
\begin{tikzpicture}[scale=1]
\begin{scope}[>={Stealth[black]},              
              every edge/.style={draw=black}]
    \node (A) at (0.5,0) {$v_1$};
    \node (B) at (2,0) {$v_2$};
    \node (C) at (0.5,1) {$v_3$};
    \node (D) at (2,1) {$v_4$};
    \node (E) at (0,2) {$v_5$};
    \node (F) at (1,2) {$v_6$};
    \node (G) at (0,3) {$v_7$};    
    \node (H) at (1.5,3) {$v_8$};   
    \path [->] (G) edge (E);
    \path [->] (H) edge (F);
    \path [->] (H) edge (D);
    \path [->] (E) edge (C);
    \path [->] (F) edge (C);
    \path [->] (C) edge (A);
    \path [->] (C) edge (B);
    \path [->] (D) edge (B);
\end{scope}
% Highlight the A_1 -> B_1 -> C_2 path. Use layers to draw
% behind everything.
\begin{pgfonlayer}{background}
	\draw[rounded corners=2em,line width=1.5em,blue!20,cap=round]
		(A.center) -- (B.east) -- (D.east) -- (F.center) -- (E.center);
\end{pgfonlayer}
\end{tikzpicture}
\caption{The Markov blanket of $v_3$ in this faithful Bayesian network is $\{v_5,v_6,v_1,v_2,v_4\}$.}
%\vskip -0.22in
\label{fg:MB}
\end{figure}
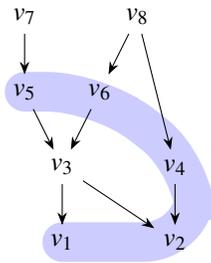

The concept of the Markov Blanket of a (target) variable is the
smallest subset of variables conditioning on which the target is
independent of the remaining variables of a
model.\footnote{Originally, this is how \citet{pearl1988probabilistic}
  defined ``Markov boundaries'', but the literature has migrated
  ``Markov blankets'' to this minimalist sense.} It tells nothing about the how the target variable is connected to the MB variables. This implies that once the MB is given, other variables beyond the blanket carry no additional information
about the target, i.e., that Markov Blankets are the optimal feature
subsets for prediction
\citep{koller1996toward,cooper1997evaluation,cheng2001kdd}. In a
faithful Bayesian network a variable's MB contains its parents and children and
its spouses (i.e., the children's other parents, see
Figure~\ref{fg:MB}). 

It follows that one approach to reducing the
complexity of learning a full causal Bayesian network is to learn the causal
structures within $n$  MBs independently, which we might call the
regional causal structure,\footnote{In view of the fact that ``local structure'' is typically used to refer to the dependencies between parameters within the conditional probability distribution of each individual variable.}
and then stitch them together. A
good review of this approach is \citet{aliferis2010locala,aliferis2010localb}.

%Although Markov Blankets are proper subsets of a domain, finding the right candidates is not much easier than learning a Bayesian network over the domain.

A natural way of doing this is suggested by the conditional
independence definition of Markov Blankets: testing dependencies
between a target and everything else given each possible subset of
other variables, looking for the minimum subset yielding zero
dependency. An exhaustive search of this kind would, of course, be
exponential in the size of the network. But we can try a heuristic,
instead of exhaustive, search. This was done by
\citet{margaritis1999bayesian}. Their work laid the foundation for a
constraint-based Markov Blanket discovery that typically consists of
alternating addition and deletion
phases. \citet{tsamardinos2003algorithms} improved on this work by
adding those variables to the candidate MB having the strongest
dependencies with the target in advance. While employing the same
statistical tests and heuristics, another algorithm that learns the
direct neighbors and spouses separately has proven superior, and hence
has been widely adopted in later constraint-based methods (see
\citet{aliferis2003hiton,tsamardinos2003time,pena2007towards,fu2008fast,aliferis2010locala,demorais2010novel,liu2016swamping}). In
particular, this improved strategy looks at the dependencies with the
target at a distance of one and two neighbors separately.  Distance
two neighbors are then filtered to remove false positives. At the end of
each learning process, the discovered MB are then required
to satisfy the symmetry condition for Markov Blankets (Proposition
\ref{prop:sym} in Section \ref{sec:mb}), which has the potential to further increase the accuracy of a learner. 

Metric-based learners, having proven themselves highly effective in
causal discovery, have subsequently been applied to Markov Blanket
discovery. In contrast to metric learning of full Bayesian networks,
the search space is restricted to local (sub-) structures around a
target variable without regard for unrelated adjacencies
\citep{cooper1997evaluation,madden2002new,acid2013score}. Given that
most real models are sparse, Markov Blankets tend to be small. This
allows exact algorithms for learning small Bayesian networks to be
applied to find optimal regional structures
independently. \citet{niinimaki2012local} published the first exact
Markov Blanket learning algorithm and also applied it to scale up to
exact general Bayesian network learning. They used dynamic programming
with the BDeu metric to find optimal local
DAGs. \citet{gao2017efficient} relaxed the symmetry enforcement in  \citet{niinimaki2012local}'s method, and proposed a method that is more efficient with similar accuracies. Based upon that method, \citet{gao2017local} developed a local-to-global Bayesian network structure learning algorithm and further reduced its computational complexity in \cite{gao2018parallel} by parallel learning regional structures. 

Other than the approaches mentioned above, Markov Blankets have also
been learned using wrapper feature selection. That is, potential MBs
are scored using predictive models such as decision trees
\citep{frey2003identifying}, linear causal models with the LASSO
estimator \citep{li2004} and ridge regularized linear models
\citep{strobl2016markov}.

%------------------------------------------------------------------------------------------------------
%------------------------------------------------------------------------------------------------------

\section{Markov Blankets} 
\label{sec:mb}

A \textit{directed acyclic graph (DAG)} is a directed graph with no directed
cycles (as in Figure~\ref{fg:MB}). We use $G=(X,E)$ to denote
a DAG over a variable set $X=\{X_1, \dots, X_n\}$ and
directed edge (arc) set $E$. We say $X_i$ is a \textit{parent} of
$X_j$ and $X_j$ is a \textit{child} of $X_i$ if there is a (directed) arc
$X_i \rightarrow X_j \in E$ from $X_i$ to $X_j$. In addition, $X_k$ is
a \textit{descendant} of $X_i$ and $X_i$ is an \textit{ancestor} of
$X_k$ if there is a directed path in $G$ from $X_i$ to $X_k$.

\begin{definition}[Markov Condition]
\label{def:markov}
Let $P$ be a joint probability distribution of the random variables in
$X$, and $G=(X,E)$ be a directed acyclic graph. We say $<G, P>$
satisfies the Markov condition if for every variable
$X_i \in X$, it is conditionally independent of its non-descendants
$ND_i$ given its parent set $\pi_i$. That is,
\begin{align*}
X_i \indep_P~ ND_i \mid \pi_i.
\end{align*} 
\end{definition}

\begin{definition}[Bayesian networks]
\label{def:bn}
Let $P$ be a joint probability distribution of the random variables in
$X$, and $G=(X,E)$ be a directed acyclic graph. We say $<G, P>$ forms
a \textit{Bayesian network} if it satisfies the Markov condition.
\end{definition}

\begin{definition}[Entailed conditional independency]
\label{def:entail}
Let $P$ be a joint probability distribution of the random variables in
$X$, and $G=(X,E)$ be a directed acyclic graph. We say $G$
\textit{entails} the conditional independency
$X_i \indep_P~ X_j \mid X_k$ for some variables $X_i, X_j, X_k \in X$, if the conditional independency holds for every joint probability distribution $P$ such that $<G, P>$ satisfies the Markov
condition.
\end{definition}

\noindent
A DAG need not entail all the conditional
independencies in a joint distribution, so the following two
definitions are introduced.

\begin{definition}[I-map]
\label{def:imap}
A directed acyclic graph $G$ is called an
\textit{independence-map (or I-map)} of a joint probability
distribution $P$, if $G$ entails all the conditional independencies in
$P$.
\end{definition}

\begin{definition}[Faithfulness]
\label{def:faithful}
A joint probability distribution $P$ is said to be \textit{faithful}
to a directed acyclic graph $G$ if $G$ entails all and only the
conditional independencies in $P$.
\end{definition}

\noindent
In this paper we assume distributions are faithful to their associated
Bayesian networks.

\begin{definition}[Markov Equivalence]
\label{def:equivalent}
Two directed acyclic graphs $G_1$ and $G_2$ are \textit{Markov equivalent} if and
only if they entail the same conditional independencies.
\end{definition}

\begin{definition}[Markov Blankets]
\label{def:mb}
Let $<G=(X,E),P>$ be a Bayesian network. The \textit{Markov Blanket}
of a variable $X_i$, denoted by $MB_i$, is the minimum subset of
$X$ such that the following holds:
\begin{align*}
X_i \indep_P ~ X\setminus \{X_i, MB_i\} \mid MB_i
\end{align*}
\end{definition}

Assuming faithfulness, being the smallest conditioning set ensures the
uniqueness of Markov Blanket \cite{tsamardinos2003towards}. Given a Bayesian network structure $G=(X,E)$, a variable $X_i$'s Markov Blanket consists
of its parents, children, and the children's other parents
(a.k.a.~spouses; see Figure~\ref{fg:MB}). We use $MB^G_i$ when we wish
to point out the DAG $G$ to which $MB_i$ belongs.

\begin{proposition}[Symmetry]
\label{prop:sym}
Let $<G=(X,E),P>$ be a Bayesian network. For any two distinct
variables $X_i, X_j \in X$:
\begin{align*}
X_j \in MB_i \Leftrightarrow X_i \in MB_j.
\end{align*}
\end{proposition}

\noindent
(See \citet{pearl1988probabilistic}, Theorem 4.)

%------------------------------------------------------------------------------------------------------
%------------------------------------------------------------------------------------------------------

\section{Minimum Message Length} 
\label{sec:mml}

Our approach to MB discovery is metric-based. In particular, we apply
the Bayesian inferential technique of Minimum Message Length (MML)
coding \citep{wallace2005}. Here we provide a brief overview of MML
and how we apply it in this research.

Minimum message length was devised by \citet{wallace1968} as a
way of balancing the complexity of a statistical model $H$ with the
fit of the model to a given data set $D$.  It implements Bayes' theorem
\begin{align*}
p(H|D) = \frac{p(H, D)}{p(D)} = \frac{p(H) \times p(D|H)}{p(D)},
\end{align*}
where $p(H)$ is the prior probability distribution of a model,
$p(D|H)$ is the likelihood of a data set given this model. In addition,
it conforms to Shannon's concept of an efficient code, satisfying
\begin{align*}
I(E) = -log(p(E))
\end{align*}
to measure the cost or information content for stating an event of
probability $p(E)$.\footnote{Throughout this paper, we use the natural
  log to calculate the MML score unless stated otherwise. Information
  is then measured in ``nits'', rather than bits.} Putting these
together, the information cost of stating a model and a data set in a
two-part message is
\begin{align}
\label{eq:mml}
I(H, D) = I(H) + I(D|H).
\end{align}
The first part $I(H)$ measures the message length for stating a model
(i.e., its structure and parameters for a certain precision). The
second part $I(D|H)$ measures how well the specified model compresses the given
data set. The aim in MML inference is to find the model having the
shortest two-part message length, and so maximizing the posterior probability
of $H$. 

A feasible approximate method for calculating the total message length
is known as \textit{MML87}, from \citet{wallace1987}. It approximates
the two parts as follows:
\begin{align}
\label{eq:mml_1}
I(H) &= -ln(p(\vec{\theta})) + \frac{1}{2} ln(F(\vec{\theta})) + \frac{|\vec{\theta}|}{2} ln(\kappa_{|\vec{\theta}|}), \\
\label{eq:mml_2}
I(D|H) &= -ln(p(D|H)) + \frac{|\vec{\theta}|}{2}.
\end{align}
For a given model with a parameter set $\vec{\theta}$ of size $|\vec{\theta}|$,
$p(\vec{\theta})$ specifies the parameter prior. The other terms in
$I(H)$ give the precision of $\vec{\theta}$, where $F(\vec{\theta})$
is the determinant of the expected Fisher information matrix and
$\kappa_{|\vec{\theta}|}$ are lattice constants
\cite{wallace2005}. The $\frac{|\vec{\theta}|}{2}$ term in $I(D|H)$ is
the extra cost of using an estimate with optimal limited
precision. (Note that a continuous datum, $d$, can only ever be
measured to limited accuracy, $\pm \frac{\epsilon}{2}$, so it has not
just a probability density, $f(d)$, but a proper probability,
$f(d)\cdot \epsilon$, assuming that the probability density function varies slowly around $d$.)

From equations (\ref{eq:mml_1}) and (\ref{eq:mml_2}), one is able to
calculate the total message length if the determinant of the expected
Fisher information matrix is calculable, and, in particular, one is
interested in knowing the MML estimates of the parameters.  Assuming
that a data set $D$ of $N$ \textit{i.i.d.} samples of a random variable comes
from a multi-state distribution, the total message length to state the
hypothesis and data set can be calculated efficiently in an alternative way by
\begin{equation}
\label{eq:msmml}
I(H, D) = \ln \left(\frac{(N+r-1)!}{(r-1)!\times \prod_{i=1}^{r} n_i!} \right).
\end{equation}
This was presented by \citet{boulton1969information} as the factorial
form of multi-state MML, where the random variable takes $r$ states and
each state appears $n_i$ times in $D$. Equation (\ref{eq:msmml}) will
be shorter than the \textit{MML87} message length by a constant
difference of $\ln\frac{\pi e}{6}$ for each parameter, because it does
not state the MML estimated parameters. The rest of this section aims at proving MML is a local consistent scoring function. 
\begin{definition}[Decomposability]
\label{def:decomp}
Let $D$ be a data set of $N$ \textit{i.i.d.} records sampled from a
Bayesian network $<G=(X,E), P>$. A metric
$I:\mathcal{G} \times \mathcal{D} \rightarrow \mathbb{R}^+$ from the set of all directed acyclic graphs and data sets over the variable set $X$ is
\textit{decomposable} if it can be written as a sum of scores for each
variable $X_i$ given its parent set $\pi_i$. That is,
\begin{align*}
I(G, D) = \sum_{X_i \in X} I(X_i|\mathbf{\pi_i}, D|_{X_i}).
\end{align*}
\end{definition}
Decomposability simplifies the calculation of a metric. For example, for
MML the second part of the message corresponds to the likelihood,
which can be factorized into a product of individual variables'
likelihood scores. Alternative metrics used in causal discovery, such
as BDe, MDL, K2, are also decomposable.

Here we make a few more assumptions. Besides faithfulness, we only consider Bayesian networks with discrete variables and no hidden variables. We assume the
parameters are independent and obey a uniform prior distribution
(which is generalized to the symmetric Dirichlet distribution in the next
section), so the parameter prior can be dealt with individually for each
variable. The next two definitions and propositions are adapted from \citet{chickering2002optimal}. I-mapness is defined structurally, without regard to parameters. Although Markov blankets, regional structures and Bayesian networks all focus on (hypothetical) structures, their scores are calculated based on an agreed parameter estimation method. Therefore, we introduce \textit{parameterized I-map}, which refers to a DAG that is an I-map of a distribution and its parameters are obtained by maximum likelihood estimation. 
 
\begin{definition}[Consistency]
\label{def:consistent}
Let $D$ be a data set of $N$ \textit{i.i.d.} records sampled from a
joint probability distribution $P$ over a variable set $X$. Assume
$G_1=(X,E_1)$ and $G_2=(X,E_2)$ are distinct directed
acyclic graphs. A metric
$I:\mathcal{G} \times \mathcal{D} \rightarrow \mathbb{R}^+$ from the set of all directed acyclic graphs and data sets over the variable set $X$ that measures
the information content for stating a model and the given data set is
\textit{consistent} if the following hold: 
\begin{enumerate}
\item if $G_1$ is an I-map of $P$ and $G_2$ is not, then $\lim\limits_{n \rightarrow \infty} I(G_1, D) < \lim\limits_{n \rightarrow \infty} I(G_2, D)$,
\item if $G_1$ and $G_2$ are both parameterized I-maps of $P$ and $G_1$ has fewer
  parameters than $G_2$, then
  $\lim\limits_{n \rightarrow \infty} I(G_1, D) < \lim\limits_{n
    \rightarrow \infty} I(G_2, D)$.
\end{enumerate}
\end{definition}

\noindent

\begin{definition}[Local Consistency]
\label{def:local_consistent}
Let $D$ be a data set of $N$ \textit{i.i.d.} records sampled from a
probability distribution $P$ over a variable set
$X$. Assume $G_1 = (X, E_1)$ and
$G_2 = (X, E_2)$ are two directed acyclic graphs such
that $E_2 = E_1 \cup \{X_i \rightarrow X_j\}$. A consistent metric
$I:\mathcal{G} \times \mathcal{D} \rightarrow \mathbb{R}^+$ from the set of all directed acyclic graphs and data sets over the variable set $X$ that
measures the information content for stating a model and the given
data set is \textit{locally consistent} if the following hold:
\begin{enumerate}
\item if $X_i \dep_P~ X_j \mid \pi_j^{G_1}$, then $\lim\limits_{n\rightarrow \infty} I(G_2, D) < \lim\limits_{n\rightarrow \infty}I(G_1, D)$,
\item if $X_i \indep_P~ X_j \mid \pi_j^{G_1}$, then $\lim\limits_{n\rightarrow \infty}I(G_1, D) < \lim\limits_{n\rightarrow \infty}I(G_2, D)$,
\end{enumerate}
where $\pi_j^{G_1}$ is the parent set of $X_j$ in $G_1$. 
\end{definition}

\begin{proposition}
Under the assumptions made above, MML is a consistent scoring function.
\end{proposition}

\begin{proof}
Since the models considered in this paper are discrete and have no hidden variables, they belong to the curved exponential family \cite{geiger2001stratified}. According to equations (\ref{eq:mml_1}) and (\ref{eq:mml_2}), the total message length can be expressed as
\begin{align*}
I(H, D) &= - \left(ln(p(D|H)) - \frac{|\vec{\theta}|}{2} a_n\right), \text{ where} \\
a_n &= 1 - \frac{2ln(p(\vec{\theta}))}{|\vec{\theta}|} + \frac{1}{|\vec{\theta}|} ln(F(\vec{\theta})) + ln(\kappa_{|\vec{\theta}|})
\end{align*}

The only term in $a_n$ that is a function of $n$ is the determinant of the expected Fisher information matrix. The likelihood grows linearly with $n$, so the determinant of the expected Fisher information matrix grows as $n^{|\theta|}$. Hence, the log of the determinant of FIM grows as $|\theta| \log n$.  Consequently, as $n \rightarrow \infty$, the term $a_n \rightarrow \infty$ slower than $n$, so $a_n/n \rightarrow 0$. From \citet{haughton1988choice}, it follows that MML must be a consistent scoring function.\footnote{Haughton's \citeyearpar{haughton1988choice} result for consistent scoring functions applies to both the linear and curved exponential families. The linear exponential family contains undirected graphical models that have no hidden variables      \citep{geiger2001stratified}. The curved exponential family contains directed acyclic graphs, chain graphs without hidden variables and several families of models (e.g., decision trees) that can approximate a full CPT. \citet{geiger2001stratified} treated graphical acyclic models with hidden variables in the stratified exponential family and emphasized that Haughton's   \citeyearpar{haughton1988choice} argument does not extend to them because some of his assumptions are violated in this family.} 
\end{proof} 

Using consistency and decomposability, one can prove that MML is a locally
consistent scoring function. This allows MML to find the optimal
Markov Blanket in the limit of infinite data.

\begin{proposition}
Under the assumptions made above, MML is a locally consistent scoring function. 
\end{proposition}

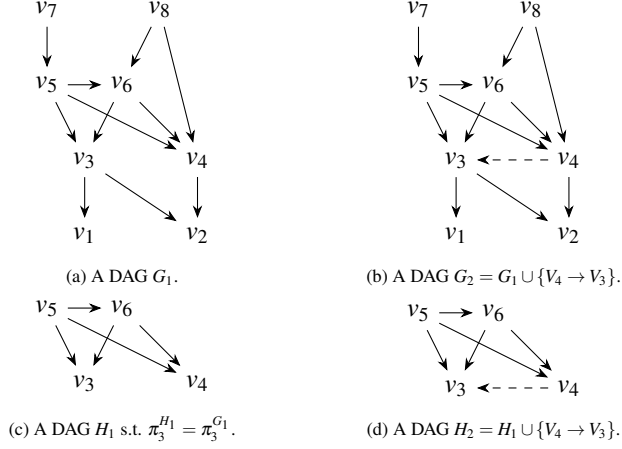
\begin{figure}[h]
\centering
\begin{subfigure}[b]{0.4\textwidth}
\centering
\begin{tikzpicture}[scale=1]
\begin{scope}[>={Stealth[black]},every edge/.style={draw=black}]
    \node (A) at (0.5,0) {$v_1$};
    \node (B) at (2,0) {$v_2$};
    \node (C) at (0.5,1) {$v_3$};
    \node (D) at (2,1) {$v_4$};
    \node (E) at (0,2) {$v_5$};
    \node (F) at (1,2) {$v_6$};
    \node (G) at (0,3) {$v_7$};    
    \node (H) at (1.5,3) {$v_8$};   
    \path [->] (G) edge (E);
    \path [->] (H) edge (F);
    \path [->] (H) edge (D);
    \path [->] (E) edge (C);
    \path [->] (E) edge (D);
    \path [->] (E) edge (F);
    \path [->] (F) edge (C);
    \path [->] (F) edge (D);
    \path [->] (C) edge (A);
    \path [->] (C) edge (B);
    \path [->] (D) edge (B);
\end{scope}
\end{tikzpicture}
\caption{A DAG $G_1$.}
\label{subfg:g1}
\end{subfigure}
\begin{subfigure}[b]{0.4\textwidth}
\centering
\begin{tikzpicture}[scale=1]
\begin{scope}[>={Stealth[black]},every edge/.style={draw=black}]
    \node (A) at (0.5,0) {$v_1$};
    \node (B) at (2,0) {$v_2$};
    \node (C) at (0.5,1) {$v_3$};
    \node (D) at (2,1) {$v_4$};
    \node (E) at (0,2) {$v_5$};
    \node (F) at (1,2) {$v_6$};
    \node (G) at (0,3) {$v_7$};    
    \node (H) at (1.5,3) {$v_8$};   
    \path [->] (G) edge (E);
    \path [->] (H) edge (F);
    \path [->] (H) edge (D);
    \path [->] (E) edge (C);
    \path [->] (E) edge (D);
    \path [->] (E) edge (F);
    \path [->] (F) edge (C);
    \path [->] (F) edge (D);
    \path [->] (C) edge (A);
    \path [->] (C) edge (B);
    \path [->] (D) edge (B);
    \path [dashed,->] (D) edge (C);
\end{scope}
\end{tikzpicture}
\caption{A DAG $G_2 = G_1 \cup \{V_4 \rightarrow V_3\}$.}
\label{subfg:g2}
\end{subfigure}

\begin{subfigure}[b]{0.4\textwidth}
\centering
\begin{tikzpicture}[scale=1]
\begin{scope}[>={Stealth[black]},every edge/.style={draw=black}]
    \node (3) at (0.5,1) {$v_3$};
    \node (4) at (2,1) {$v_4$};
    \node (5) at (0,2) {$v_5$};
    \node (6) at (1,2) {$v_6$};
    \path [->] (6) edge (3);
    \path [->] (5) edge (3);
    \path [->] (5) edge (6);
    \path [->] (5) edge (4);
    \path [->] (6) edge (4);
\end{scope}
\end{tikzpicture}
\caption{A DAG $H_1$ s.t. $\pi_3^{H_1}=\pi_3^{G_1}$.}
\label{subfg:h1}
\end{subfigure}
\begin{subfigure}[b]{0.4\textwidth}
\centering
\begin{tikzpicture}[scale=1]
\begin{scope}[>={Stealth[black]},every edge/.style={draw=black}]
    \node (3) at (0.5,1) {$v_3$};
    \node (4) at (2,1) {$v_4$};
    \node (5) at (0,2) {$v_5$};
    \node (6) at (1,2) {$v_6$};
    \path [->] (6) edge (3);
    \path [->] (5) edge (3);
    \path [->] (5) edge (6);
    \path [->] (5) edge (4);
    \path [->] (6) edge (4);
    \path [dashed,->] (4) edge (3);
\end{scope}
\end{tikzpicture}
\caption{A DAG $H_2 = H_1 \cup \{V_4 \rightarrow V_3\}$.}
\label{subfg:h2}
\end{subfigure}
\caption{The score change between $H_1$ and $H_2$ is identical to the score change between $G_1$ and $G_2$ because of the decomposibility of MML.}
\label{fg:mml is local consistent}
\end{figure}

\begin{proof}
Let $D$ be a set of \textit{i.i.d.} samples generated from a distribution $P$ over a variable set $X$. Let $G_1=(X,E_1)$ and $G_2=(X,E_1 \cup \{X_i\rightarrow X_j\})$ be two DAGs different by exactly one edge, e.g., as shown in Figure \ref{subfg:g1} and \ref{subfg:g2}. Then there is a pair of DAGs $H_1=(Y, F_1)$ and $H_2=(Y, F_1 \cup \{X_i\rightarrow X_j\})$ over a subset $Y \subseteq X$ of variables such that in $H_1$ the parent set for the variable $X_j$ satisfies $\pi_j^{H_1} = \pi_j^{G_1}$ and $H_2$ is a complete DAG, such as shown in Figure \ref{subfg:h1} and \ref{subfg:h2}. If $X_i \dep_P~ X_j \mid \pi_j^{G_1}$ then $X_i \dep_{P|_Y}~ X_j \mid \pi_j^{H_1}$, so $H_2$ is an I-map of the joint distribution $P|_{Y}$ restricted to the variable subset $Y$ whilst $H_1$ is not. By decomposability and consistency of MML we have $\lim\limits_{n\rightarrow \infty}I(G_1, D) - \lim\limits_{n\rightarrow \infty} I(G_2, D) = \lim\limits_{n\rightarrow \infty}I(H_1, D|_{\mathbf{Y}}) - \lim\limits_{n\rightarrow \infty} I(H_2, D|_{\mathbf{Y}}) = d > 0$.

On the other hand, if $X_i \indep_P~ X_j \mid \pi_j^{G_1}$, then both $H_1$ and $H_2$ are parameterized I-maps of $P|_{Y}$ but the former has fewer parameters than the latter. Hence, $\lim\limits_{n\rightarrow \infty}I(G_1, D) - \lim\limits_{n\rightarrow \infty} I(G_2, D) =  \lim\limits_{n\rightarrow \infty}I(H_1, D|_{\mathbf{Y}}) - \lim\limits_{n\rightarrow \infty} I(H_2, D|_{\mathbf{Y}}) = d < 0$.
\end{proof}

% this is chickering's proof
\iffalse
\begin{proof}
  The proof follows from the fact that in the limit, the criterion
  ranks models in the same order as BIC. Because BIC is decomposable,
  the increase in score that results from adding the edge $Xi
  \rightarrow Xj$ to
  any DAG G is the same as the increase in score that results from
  adding the edge to any other DAG H for which Xj has the same
  parents. We can therefore choose a particular H---where PaHj = PaGj
  ---for which adding the edge $Xi \rightarrow Xj$ results in a complete DAG H′;
  that is, H′ has an edge between every pair of nodes. Because the
  complete DAG imposes no constraints on the joint distribution, the
  lemma follows immediately from the consistency of BIC.
\end{proof}
\fi

\noindent
It is worth noting that MML differentiates between DAGs in the same
Markov equivalence class, but only to the extent of a prior inductive
bias favoring simpler models.

%------------------------------------------------------------------------------------------------------
%------------------------------------------------------------------------------------------------------

\section{Learning Markov Blankets using MML}
\label{sec:mbmml}

The problem we set ourselves was to search the space of Markov
Blankets for each variable in a data set to find a complete set of MBs
that minimizes an MML score (equivalently, maximizes the corresponding
posterior Bayesian score). Of course, in principle this involves
searching the exponential space of all possible subsets of variables,
so we used a heuristic greedy search rather than exhaustive
search. For MML to operate, we also had to define a model space for
representing the probability distribution of each target variable
given its Markov Blanket. The ideal model structure would be the subgraph of the true DAG induced by the Markov Blanket, on the general principle that you can't outdo the truth. But since we
don't know the true causal DAG, we tried a variety of models which can
plausibly do a good job of representing that conditional probability
distribution: a conditional probability table (CPT) reflecting all MB
variables as parents of the target, which maximizes the number of
parameters, meaning it has maximal representational power at the
expense of requiring the most data to parameterize accurately; a Naive
Bayes (NB) model that assumes independence between all MB variables
given the value of the target variable, which minimizes the number of
parameters at the expense of misrepresenting dependencies between
them; and Markov Blanket polytrees (MBPs), which compromise between
these two extremes by representing MB variables as related to the
target variable and other MB variables via a singly connected
DAG. There are many other alternative local models discussed in the
broader literature (e.g., \citet{neil1999learning}), but here we limit
ourselves to these three.

We now explain each of these models and their MML scores in detail.

\subsection{MML for CPT models}
\label{sec:mml_cpt}

For any discrete variable $X_i \in X$, its probability
density function conditioning on the full joint distribution of its
parents set $\pi_i$ can be expressed by a $r_i \times r_{\pi_i}$
conditional probability table (CPT), where $r_i$ and $r_{\pi_i}$ are
the number of states of $X_i$ and $\pi_i$ respectively (while the
densities of continuous variables can be approximated by such a
table). We use a CPT model to describe the relation between a target
and its MB variables by treating those variables as if they are all
parents, without claiming they actually are all parents, much as in a
multiple regression model.  A full CPT can capture any interactions
between the MB variables (e.g., an XOR) as long as there are enough
data to support effective parameterization; this is a requirement that
grows exponentially in $r_{\pi_i}$. We use $\phi_i(S)$ to denote the
CPT model of $X_i$ with a subset $S \subseteq X$ being the
hypothetical parent set of $X_i$.

The parent instantiations partition $X_i$ into $r_s$ multi-state
distributions. By the parameter independence assumption, the message
length of a CPT model is a sum of the message length of each
multi-state distribution over all $r_s$ partitions. Assuming the
parameters follow Dirichlet distributions, multi-state MML
can be applied. Hence, the total message length for stating a CPT
model $\phi_i(S)$ with the hypothetical parent set $S$ and the data $D_{X_i}$ for the target $X_i$ is
\begin{align}
\label{eq:mmlcpt}
I(\phi_i(S), D_{X_i}) = & \sum_{j = 1}^{r_s} \ln \left(\frac{(n_j+\alpha_0-1)! \prod_{k=1}^{r_i} (\alpha_k - 1)!}{(\alpha_0-1)! \prod_{k=1}^{r_i} (n_{jk} + \alpha_k - 1)!} \right) + \frac{r_s(r_i-1)}{2} \ln \frac{\pi e}{6},
\end{align}
where $\vec{\alpha} = (\alpha_1, \dots, \alpha_{r_i})$ is the vector of Dirichlet's concentration parameters for the variable $X_i$. The parameter value of each of $X_i$'s state is controlled by the corresponding $\alpha_k \in \vec{\alpha}$ and define $\alpha_0 = \sum \vec{\alpha}$. The term $n_{jk}$ is the count of matching data points for $\pi_i$ being
in state $j$ and $X_i$ in state $k$, and
$n_j = \sum_{k=1}^{r_i} n_{jk}$. In the tested models we have no prior knowledge favoring one state over another, so we used a symmetric Dirichlet distribution with a concentration parameter $1$ for all $r_i$ states of a variable $X_i$. This is the same as sampling from a uniform distribution over all states, so the above equation generalizes equation \ref{eq:msmml}.

Suppose only a learned CPT model is used to encode a data set, the next proposition shows that the shortest MML code length in the limit  is achieved when the hypothetical parent set of $X_i$ equals $MB_i$.
\begin{proposition}
\label{prop:mmlcpt}
Let $D$ be a data set with $N$ \textit{i.i.d.} records sampled from a
joint probability distribution $P$ over variables
$X=\{X_1, \dots, X_n\}$. The MML score for stating a CPT
model of $X_i$ and the given data set satisfies the following:
\begin{align*}
\lim\limits_{n\rightarrow \infty}I(\phi_i(MB_i), D_{X_i}) <
  \lim\limits_{n\rightarrow \infty}I(\phi_i(S), D_{X_i}), \forall S
  \subseteq X \text{ s.t. } S \neq MB_i.
\end{align*}
\end{proposition}

\begin{proof}
Suppose the subset $S = MB_i \cup \{X_j\}$ such that the variable $X_j \notin MB_i$. Let $G_1$ be a DAG over $\{X_i, X_j\}\cup MB_i$ such that the parent sets $\pi_i^{G_1} = MB_i$ and $\pi_k^{G_1} = \emptyset$ for all $X_k \in \{X_j\}\cup MB_i$. In addition, let $G_2$ be the same as $G_1$ but with an additional edge $X_j \rightarrow X_i$. Since $X_j \notin MB_i$, we have $X_i \indep_P X_j | \pi_i^{G_1}$. By the local consistency of MML, the scores of the two models satisfy $\lim\limits_{n \rightarrow \infty}I(G_1, D_{G_1}) < \lim\limits_{n \rightarrow \infty}I(G_2, D_{G_2})$. Since all variables in $G_1, G_2$ have the same parent sets except for $X_i$ and MML is a decomposable scoring function, we have  $\lim\limits_{n\rightarrow \infty}I(\phi_i(MB_i), D_{X_i}) = \lim\limits_{n\rightarrow \infty}I(X_i|MB_i, D_{X_i}) < \lim\limits_{n\rightarrow \infty}I(X_i|S, D_{X_i}) = \lim\limits_{n\rightarrow \infty}I(\phi_i(S), D_{X_i})$.

Suppose the subset $S = MB_i \setminus \{X_j\}$. Similarly, define $G_1, G_2$ as above but $\pi_i^{G_1} = S$. Since $X_j \in MB_i$, it implies that $X_i \dep_P X_j | \pi_i^{G_1}$. Then local consistency implies $\lim\limits_{n \rightarrow \infty}I(G_2, D_{G_2}) < \lim\limits_{n \rightarrow \infty}I(G_1, D_{G_1})$. For the same reasons, we have $\lim\limits_{n\rightarrow \infty}I(\phi_i(MB_i), D_{X_i}) < \lim\limits_{n\rightarrow \infty}I(\phi_i(S), D_{X_i})$. 
\end{proof}

\subsection{MML for Naive Bayes models}

Naive Bayes (NB) models invert the structure of regressions: a central
(target) variable is the parent of all the attributes, inducing a
marginal dependency between every pair (if faithful), while inducing a
conditional independency between them. It is very popular in machine
learning for two reasons: it minimizes the number of parameters,
making it useful even in data poor environments; it works reasonably
well on many problems, even many that violate the independence
assumption so long as the dependencies omitted are not overly strong.
With the conditional independence assumption,
Naive Bayes parameters increase linearly in the number of variables,
which makes it useful for dealing with large problems. %Another benefit of using an NB model is that the joint density $p(X_i, S)$ factorizes into a simple product of conditional probabilities.
The posterior probability of the target variable $X_i$ given a hypothetical child set $S$ is
\begin{align}
p(X_i|S) = \frac{p(X_i)\prod_{X_j \in S} p(X_j|X_i)}{\sum_{x_i=1}^{r_i} p(x_i)\prod_{X_j \in S} p(X_j|x_i)},
\end{align}
where $p(x_i)$ is a short for $p(X_i = x_i)$. Each term $p(X_i)$ and $p(X_j | X_i)$
is a single or a set of multi-state distributions, so can be calculated using adaptive coding MML by Equation \ref{eq:mmlcpt} or Equation \ref{eq:msmml} if priors are assumed to come from a symmetric Dirichlet distribution. Hence, the total
message length for stating an NB model and the data for the target is
\begin{align}
\label{eq:mmlnb}
\begin{split}
I(\phi_i(S), D_{X_i}) = -\ln p(X_i) - \sum_{X_j \in S} \ln p(X_j|X_i) 
+ \ln \sum_{x_i=1}^{r_i} p(x_i)\prod_{X_j \in S} p(X_j|x_i).
\end{split}
\end{align}
Notice that this is the message length omitting the MML estimate of the parameters. 

\subsection{MML for Markov Blanket polytrees}

Between the extremes of a regression structure, with all attributes as
independent parents of the target, and NB models, with all attributes
as isolated children, come almost every other possible DAG structure
relating MB variables with their target. The true model is likely to
be amongst them, but as with many learning problems where the truth is
unknown, some ensembling approach suggests itself as a way of
approximating the truth. Here we use an ensembling method that samples
as many local polytrees as possible, then outputs a
weighted average message length over all the samples.  This way the MB
variables with a good variety of network structures, allowing many
interactions to be modeled, but is nevertheless limited, so that
the number of model parameters on average is less than that of local DAG's.

We call the restricted local structures being sampled Markov Blanket
polytrees (MBPs). A polytree is a DAG such that its underlying
undirected graph is a tree.
\begin{definition}
Let $<G=(X,E),P>$ be a Bayesian network. A \textit{Markov Blanket polytree} $T_i$ of a target variable $X_i$ is a polytree over the variables $\{X_i\} \cup MB_i$ such that 
\begin{align*}
MB^{T_i}(X_i) = MB^G(X_i).
\end{align*}
\end{definition}

%Searching through the space of MBPs allows a variable to be tested under different roles of a Markov Blanket candidate. Having multiple paths between pair of nodes, a DAG could be a supergraph of more than one polytree over the same variables, all of which are true subgraphs of the DAG. Hence,  

The next proposition presents a recursive formula for counting the
number of labeled Markov Blanket polytrees (MBPs) over a set of $n$
variables.
\begin{proposition}
\label{prop:nmbps}
Let $Y$ be a variable whose Markov Blanket contains
$n \in [1, \infty)$ variables. The number $f(n)$ of labeled Markov Blanket
polytrees for $Y$ can be computed by the following recursive equation
\begin{align}
\label{eq:nmbps}
f(n) &= \sum_{i=0}^n \binom{n}{i} + \sum_{m=1}^{\floor*{\frac{n}{2}}} ~\sum_{k=1}^{n-2m+1} g(n,m,k), \\ \nonumber
g(n,m,k) &= \binom{n}{k+1}(k+1) \sum_{k'=1}^{\min\{k, n-k-2(m-1)\}}\frac{q}{m} \cdot g(n-k-1,m-1,k'),
\end{align}
where $q=1$ if $k=k'$ and $m$ otherwise.
\end{proposition}

\begin{proof}
It is trivial to bound the number of colliders $m \in \left[0, \floor*{\frac{n}{2}}\right]$. \\
\underline{Case 1:} When $m=0$\\
$MB_i$ contains only parents and$/$or children. There are $\binom{n}{i}$ ways of selecting $i \in [0, n]$ children from $n$ labeled nodes. The order of these parents or children does not matter in a polytree. Therefore, the number of labeled MBPs when $m=0$ is 
\begin{equation}
\label{enum_m0} 
\sum_{i=0}^n \binom{n}{i}.
\end{equation}
\underline{Case 2:} When $m > 0$\\
Each of $Y$'s children and its spouses (if there are any) forms a branch. The largest branch with $k$ spouses can be enumerated in 
\begin{equation}
\label{enum_m1_branch}
\binom{n}{k+1}(k+1)
\end{equation}  
ways, where $k \in [1, n-2m+1]$. There are ${n \choose k + 1}$ ways
selecting $k+1$ nodes to form the largest branch. And each one of the
$k+1$ nodes needs to be a common child once in order to fully enumerate all
cases. $k$'s upper bound is obtained if each of the other $m-1$
branches contains only a collider and a spouse, in which case
$n - 2(m-1) -1 = n-2m+1$. Hence, when $m > 0$ the number of MBPs can
be obtained by multiplying equation (\ref{enum_m1_branch}) with the
total enumeration of the remaining $n-k-1$ nodes. The subgraph over
the remaining nodes can be counted by the same approach. By doing this
recursively, we will end up with a subgraph in which $Y$ has no
spouse. It can then be enumerated by equation
(\ref{enum_m0}). Therefore, the total enumeration of MBPs when $m > 0$
is
\begin{equation}
\label{enum_m1}
\sum_{m=1}^{\floor*{\frac{n}{2}}} \sum_{k=1}^{n-2m+1} g(n,m,k), 
\end{equation}
for 
\begin{align}
\label{enum_gnmk}
%\begin{split}
g(n,m,k) = \binom{n}{k+1}(k+1) \sum_{k'=1}^{\min\{k, n-k-2(m-1)\}}\frac{q}{m} \cdot g(n-k-1,m-1,k'),
%\end{split}
\end{align}
with $q=1$ if $k=k'$ and $m$ otherwise. The maximum number of spouses
$k'$ in a subgraph is bounded above by the minimum between the maximum
number of available nodes $n-k-2(m-1)$ and $k$ from its supergraph.

\begin{figure}[h]
\centering
\begin{subfigure}[b]{0.45\textwidth}
\centering
\begin{tikzpicture}[scale=1]
\begin{scope}[>={Stealth[black]},every edge/.style={draw=black}]
    \node (1) at (0,0) {$v_1$};
    \node (2) at (-1,0) {$v_2$};
    \node (3) at (-0.5,-1) {$v_3$};
    \node (4) at (1,0) {$v_4$};
    \node (5) at (0.5,-1) {$v_5$};
    \path [->] (1) edge (3);
    \path [->] (2) edge (3);
    \path [->] (1) edge (5);
    \path [->] (4) edge (5);
\end{scope}
\draw[blue,thick,dotted] ($(2.north west)+(-0.2,0.2)$)  rectangle ($(3.south east)+(0.2,-0.2)$);
\end{tikzpicture}
\caption{Start with labels $V_2,V_3$.}
\label{subfg:enum_dup_a}
\end{subfigure}
\begin{subfigure}[b]{0.45\textwidth}
\centering
\begin{tikzpicture}[scale=1]
\begin{scope}[>={Stealth[black]},every edge/.style={draw=black}]
    \node (1) at (0,0) {$v_1$};
    \node (4) at (-1,0) {$v_4$};
    \node (5) at (-0.5,-1) {$v_5$};
    \node (2) at (1,0) {$v_2$};
    \node (3) at (0.5,-1) {$v_3$};
    \path [->] (1) edge (3);
    \path [->] (2) edge (3);
    \path [->] (1) edge (5);
    \path [->] (4) edge (5);
\end{scope}
\draw[blue,thick,dotted] ($(4.north west)+(-0.2,0.2)$)  rectangle ($(5.south east)+(0.2,-0.2)$);
\end{tikzpicture}
\caption{Start with labels $V_4,V_5$.}
\label{subfg:enum_dup_b}
\end{subfigure}
\caption{Double counting the same MBP for the case of two colliders of the same size.}
\label{fg:double counting mbpt}
\end{figure}
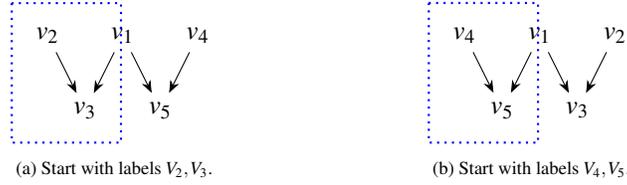

As the largest branch is enumerated independently from the remaining
nodes, some of the graphs are counted multiple times. For example, to enumerate the MBPs for the variable $V_1$ with $n=4$ MB candidates, when going through the case where there are two colliders (i.e., $m=2$),  we obtain Figure \ref{subfg:enum_dup_a} when labelling the largest branch (i.e., left/right) with $\{V2, V3\}$, and Figure \ref{subfg:enum_dup_b} when
labelling the largest branch (i.e., left/right) with $\{V4, V5\}$. The
resulting two labeled graphs, however, are identical, and hence we
divide the total number by $\frac{1}{2}$. In general, the total
number needs to be divided by $\frac{1}{m}$; hence
$\frac{q}{m}$ appears in equation (\ref{enum_gnmk}).  
\end{proof}

The total number of Markov Blanket polytrees (MBPs) is dramatically
reduced compared with DAGs, as shown in Table \ref{tb:nmbps}.

\begin{table}[h]
\centering
\caption{The number of labeled DAGs and MBPs on $n \in [1, 7]$ nodes (i.e., [0,6] MB candidates).}
\label{tb:nmbps}
\begin{tabular}{lll}
\hline
\# nodes & \# DAGs    & \# MBPTs \\ \hline
1        & 1          & 1        \\
2        & 3          & 2        \\
3        & 25         & 6        \\
4        & 543        & 23       \\
5        & 29281      & 104      \\
6        & 3781503    & 537      \\
7        & 1138779265 & 3100     \\ \hline
\end{tabular}
\end{table}

The message length for transmitting data using an MBP model is
calculated as the log of the conditional probability
\begin{align*}
p(X_i|S) = \frac{p(X_i \mid \pi_i^{T_i})\prod_{X_j \in S} p(X_j | \pi_j^{T_i})}{\sum_{x_i=1}^{r_i}p(x_i \mid \pi_i^{T_i})\prod_{X_j \in S} p(X_j | \pi_j^{T_i})}
\end{align*}
which is factorized into a product of each variable's probability
conditioned on its parent set in a Markov Blanket polytree $T_i$ as
estimated using the adaptive code method. Hence, the total message
length has the form
\begin{align}
\label{eq:mmlmbp}
\begin{split}
I(\phi_i(S), D_{X_i}) = -\ln p(X_i\mid \pi_i^{T_i}) - \sum_{X_j \in S} \ln p(X_j | \pi_j^{T_i}) + \ln \sum_{x_i=1}^{r_i}p(x_i \mid \pi_i^{T_i})\prod_{X_j \in S} p(X_j | \pi_j^{T_i}).
\end{split}
\end{align}
To calculate a weighted average score, we uniformly average the
conditional probabilities $p(X_i\mid S)$ over all sampled MBPs
in the set $\mathcal{T}_i$ containing the variables $\{X_i\} \cup MB_i$, then
take the negative $\log$. The uniform prior can be replaced by any
reasonable prior over the possible Markov Blanket polytrees.

\subsection{The MBMML algorithm}
This section presents two MBMML algorithms in pseudocode for learning
Markov Blankets using either a fixed local structure (i.e., CPT or NB)
or an ensemble of random local structures (i.e., MBPs).\footnote{In
  this work, when we refer to randomly generated DAGs these refer to
  networks generated by: uniformly selecting a total ordering of
  variables; then for each node in order, uniformly select the number
  of its parents up to min(\# of predecessors, the maximum fan-in for
  the experiment); uniformly select its parents from amongst its
  predecessors. In the case of MBPs in particular, any arc addition is
  suppressed if it would introduce an undirected loop.} Both
algorithms use greedy search starting with an empty Markov
Blanket and iteratively adding the highest ranked candidate for
reducing the total message length (equation \ref{eq:mmlcpt} or
\ref{eq:mmlnb}, or \ref{eq:mmlmbp}). Both algorithms stop and output a
learned Markov Blanket if no scores can be improved by adding a single
variable to the current MB.

\begin{algorithm}[]
\caption{MB discovery using MBMML+CPT/NB}
\label{alg:mbmmlf}
\begin{algorithmic}[MBMML]
\Procedure{$MBMML$}{$D, X, X_i, \phi_i$}, where $D$ is a given data set, $X$ is the set of all variables, $X_i$ is the target variable and $\phi_i$ is fixed to be either a CPT or  NB model.
    \State $S = X \setminus {X_i}$ \Comment{unchecked variables}
    \State $Z = \emptyset$ \Comment{learned MB}
    \State $L = I(\phi_i(\emptyset), D_{X_i})$  \Comment{empty model score}
    \While {$S \neq \emptyset$}
    		\State $X_k = \argmin_{X_j} I(\phi_i(Z \cup \{X_j\}), D_{X_i}), \forall X_j \in S$ \Comment{best candidate}
    		\State $L' = I(\phi_i(Z \cup \{X_k\}), D_{X_i})$ \Comment{current best score}
    		\If{$L' < L$} \Comment{admit when score reduces}
    			\State $Z = Z \cup \{X_k\}$
    			\State $S = S \setminus \{X_k\}$
    			\State $L = L'$ \Comment{update best score}
    		\Else
    			\State Stop 
			\EndIf
	\EndWhile
	\State Output $Z$
\EndProcedure
\end{algorithmic}
\end{algorithm}

% Insert the algorithm
\begin{algorithm}[]
\caption{MB discovery using MBMML+MBP}
\label{alg:mbmmlr}
\begin{algorithmic}[MBMML]
\Procedure{$MBMML$}{$D, X, X_i, \phi_i, K$}, where $D$ is a given data set, $X$ is the set of all variables, $X_i$ is the target variable, $\phi_i$ is a MBP model, $K$ is the number of randomly sampled MBPs. 
    \State $S = X \setminus {X_i}$ \Comment{unchecked variables}
    \State $Z = \emptyset$ \Comment{learned MB}
    \State $L = I(\phi_i(\emptyset), D_{X_i})$  \Comment{empty model score}
    \While {$S \neq \emptyset$}
    		\If{$f(|Z| + 1) \leq K$} \Comment{number of MBPs by equation \ref{eq:nmbps}}
    			\State $\mathcal{T}_i := \{\text{all MBPs over } Z \cup \{X_j\}\}$ \Comment{all MBPs}
    		\Else
    			\State $\mathcal{T}_i =  \{K \text{ random MBPs over } Z \cup \{X_j\}\}$ \Comment{randomly sampled MBPs}
    		\EndIf 
    		\State $X_k = \argmin_{X_j} E_{\mathcal{T}_i}(I(\phi_i(Z \cup \{X_j\}), D_{X_i})), \forall X_j \in S$ \Comment{best candidate}
    		\State $L' = E_{\mathcal{T}_i}(I(\phi_i(Z \cup \{X_k\}), D_{X_i}))$ \Comment{current best expected score}
    		\If{$L' < L$} \Comment{admit when score reduces}
    			\State $Z = Z \cup \{X_k\}$
    			\State $S = S \setminus \{X_k\}$
    			\State $L = L'$ \Comment{update best score}
    		\Else
    			\State Stop 
			\EndIf
	\EndWhile
	\State Output $Z$
\EndProcedure
\end{algorithmic}
\end{algorithm}

To ensure there is no conflict between the learned Markov Blankets, we
pass the outputs from both MBMML+CPT/NB and MBMML+MBP algorithms
to a symmetry enforcement algorithm, per \S \ref{sec:mb}. There are
two simple ways of enforcing symmetry, taking either the union or the
intersection of two neighboring Markov Blankets. We used Union
enforcement for MBMML+CPT, because a CPT model's precision converges
to $1$ as sample size increases. So, its exponential increase in
parameters is likely to result in more false negatives than false
positives. We used Intersection enforcement for MBMML+NB, because a
Naive Bayes model will produce more false positives than a CPT model
due to its lack of representational power, but fewer false
negatives. It is unclear which enforcement is a better option for
MBMML+MBP, but we took Union enforcement. %\textcolor{red}{We didn't apply symmetry enforcement on PCMB and SLL because SLL ensures symmetry during learning and PCMB also enforce symmetry(?) but the results after enforcing symmetry don't change much. But IAMB with symmetry under the union rule has dramatic improvement in accuracy, it is in fact better than mmlcpt in most cases. Be aware of how to deal with this if referee asks!} \footnote{A recent publication by \citet{zhao2019improving} proposed score-based and constraint-based methods to enforce symmetry. That work appears to have the potential to increase the accuracy of the MML methods presented here but is not in the scope of the current work.}

The process of the symmetry enforcement is shown in Algorithm
\ref{alg:sym}.

\begin{algorithm}[]
\caption{Symmetry enforcement}
\label{alg:sym}
\begin{algorithmic}[]
\Procedure{}{} Given the learned Markov Blankets $\{MB_i\}, \forall X_i \in X$
\For{each $MB_i$}
	\For{each $X_j \in MB_i$}
		\If{$X_i \notin MB_j$}
			\If{Union}
				\State $MB_j = MB_j \cup \{X_i\}$
			\Else{ Intersection}
				\State $MB_i = MB_i \setminus \{X_j\}$
			\EndIf
		\EndIf
	\EndFor
\EndFor
	\State Output $\{MB_i\}$
\EndProcedure
\end{algorithmic}
\end{algorithm}

%------------------------------------------------------------------------------------------------------
%------------------------------------------------------------------------------------------------------

\section{Experiments with Markov Blanket discovery}
\label{sec:experiments}

In this section, we present our experimental results testing our three
MB discovery methods, MBMML+CPT, MBMML+NB, and MBMML+MBP, against the
three leading alternatives:
\begin{itemize}
\item IAMB (Iterative Associative Markov Blanket learning) is a
  constraint-based algorithm that uses a mutual information (MI) test for candidate admission \citep{tsamardinos2003algorithms}. In phase 1 it
  adds variables that are interdependent with the target variable when
  conditioning on the current candidate Markov Blanket. That is
  followed by a pruning phase, deleting any variables found to be
  false positives.
\item PCMB (Parent-Children Markov Blanket learning) is another two-phase
  constraint-based algorithm \citep{pena2007towards}. First, it finds the direct
  neighbors of the target using a conditional independence test ($G^2$ test). It then finds the neighbors of each neighbor of the target, and prunes away false positives. 
\item SLL (Score-based Local Learning), by contrast with earlier MB
  learners, is metric-based, using the BDeu score in a dynamic
  programming algorithm \citep{niinimaki2012local}. It is an exact
  algorithm that searches through the entire space of equivalence
  classes of local DAGs around the target variable, then reads off the
  Markov Blanket from the optimal DAG. \citet{niinimaki2012local} used
  SLL also to scale up general Bayesian network structure
  learning. 
\end{itemize}
We used the implementation of PCMB provided by
\citet{pena2007towards} and IAMB in R's \textit{bnlearn} package \cite{scutari2009learning}. We set the significance level $\alpha=0.01$ for both algorithms for conditional independence tests as what used in \cite{pena2007towards}. We used SLL's source code from
\citet{niinimaki2012local} and its default equivalent sample size $1$
for BDeu. SLL reverts to the GES algorithm
\cite{chickering2002learning} for learning the MB's local structure if
it finds more than $20$ variables in the MB. The three MML methods
assumed uniform parameter prior (i.e., symmetric Dirichlet with
a signle concentration parameter $\alpha=1$). The MBMML+MBP algorithm
was set to randomly sample $100$ Markov Blanket polytrees from the
total polytree space and took the unweighted average of their message
lengths.
%The IPCMB algorithm has been experimentally justified to outperform IAMB and PCMB. And the IPCMB recently has been justified to be less superior than SLL and STMB \cite{gao2017efficient} on similar data sets. 

\begin{table}[hbt]
\caption{Summary of tested Bayesian networks. 30-5-4-1 refers to artificial networks with 30 variables, maximum fan-in 5, maximum number of states 4 per variable, and uniform ($\vec{\alpha}=<1>$) parameter prior.}
\label{tab:exp_models}       
\begin{tabular}{llll}
\hline\noalign{\smallskip}
Network    & \# variables & Max fan-in & Mean MB size  \\
\noalign{\smallskip}\hline\noalign{\smallskip}
CHILD      & 20    & 2  &  3 \\
INSURANCE & 27    & 3  &  5.19 \\
ALARM     & 37   & 4  & 3.51 \\  
BARLEY & 48 & 4 & 5.25 \\
HAILFINDER & 56 & 4 & 3.54 \\
30-5-4-1 & 30 & 5 & 8 \\
50-5-4-1 & 50 & 5 & 9.73 \\
80-5-4-1 & 80 & 5 & 10.08 \\
\noalign{\smallskip}\hline
\end{tabular}
\end{table}

Section \ref{sec:exp_real} focuses on testing with real models (Table
\ref{tab:exp_models}) and test data sets provided by
\citet{tsamardinos2006}, which we identify by name.
\footnote{\url{http://pages.mtu.edu/~lebrown/supplements/mmhc_paper/mmhc_index.html}}
These models and data sets have often been used for testing Markov
Blanket and causal discovery learners. 
\footnote{These are presumptively the true models, but were in fact built by the researchers, and hence are not guaranteed to be the true generative models.}
Section \ref{sec:exp_artificial} then extends the experiments to artificial
Bayesian networks containing $30$ or $50$ variables, with fixed maximum fan-ins and arities. For each class of model, we randomly generated $5$ different Bayesian networks, each
of which was then used to generate $5$ different data sets for each of
the sample sizes 100, 500, 2000, 5000.

We used precision, recall, and edit distance to evaluate the
performance of the different algorithms in finding the true MBs, that
is, the percentage of true positives amongst those asserted to be in
the MB (true plus false positives), the percentage of MB variables
that were found (i.e., the ratio TP: TP+FN), and the sum of false
positives and false negatives, respectively. These are all
accuracy-oriented measures, which in a strictly methodological
(non-applied) study may be about the best we can expect to do. That
is, utilities for different kinds of errors or successes can't be
stated in the abstract, so Bayesian evaluation measures are hard to
identify. Nevertheless, we might assert some preference for precision
over recall, on the grounds that our intended purpose is to improve
causal discovery, for example by feeding the results of MB discovery
into a causal discovery process. In that domain, it seems at least
plausible that false positives (reducing precision) are more damaging
than false negatives (reducing recall), since falsely asserting
membership in an MB would positively mislead subsequent causal
discovery, whereas an error of omission leaves causal discovery no
worse off than not doing any MB discovery, with respect to that
variable and MB, at any rate. We don't have the temerity to try to
quantify that intuition, however we shall consider it in interpreting
our results. Edit distance is a kind of compromise between these two
accuracy measures.

Those results will all be reported using $95\%$ confidence intervals,
in line with current APA guidelines \citep{apa2013}, either in the
main text or appendices.

\subsection{Accuracy on real models}
\label{sec:exp_real}

\begin{figure}[h]
  \centering
    \includegraphics[scale=0.6]{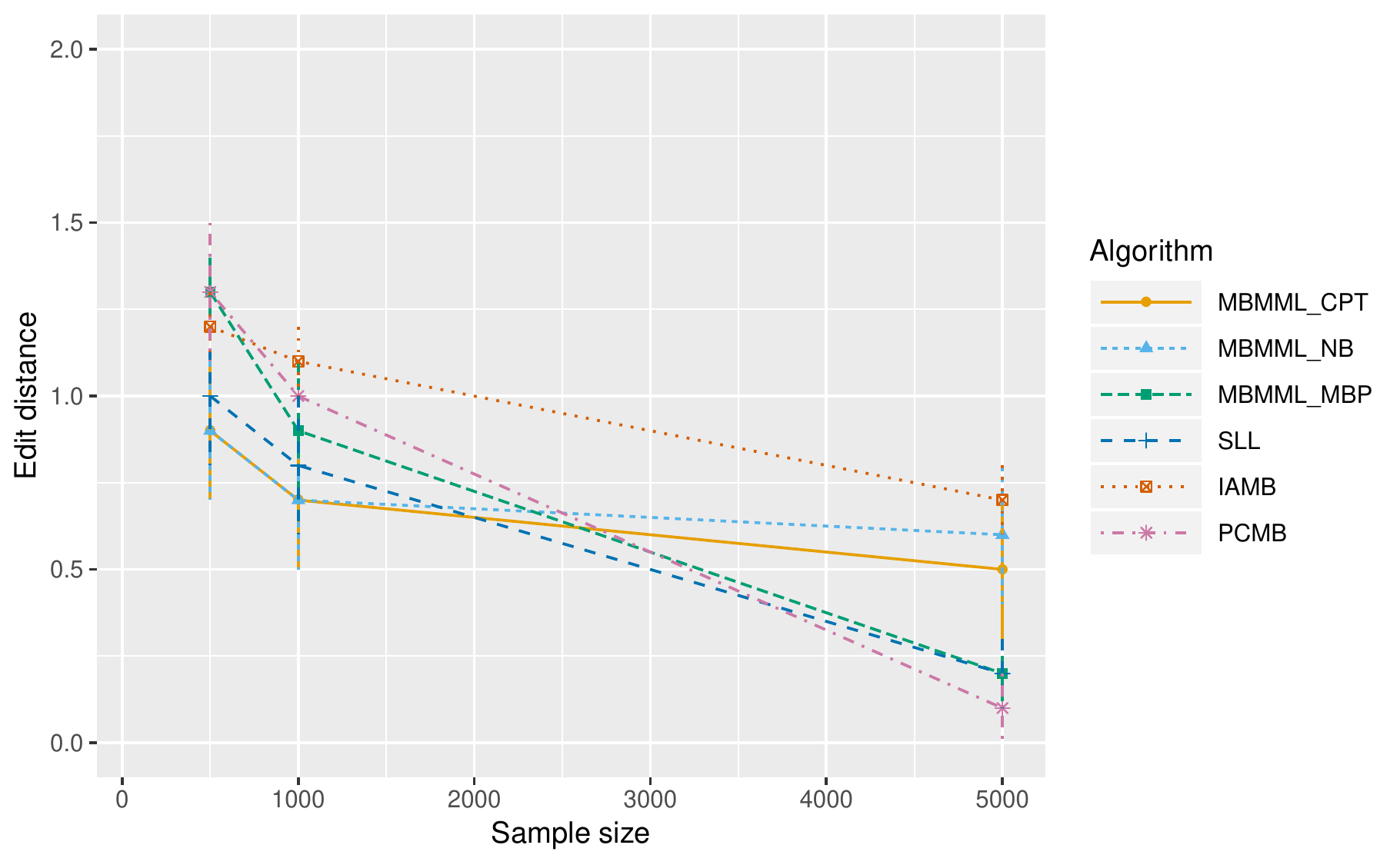}
  \caption{Edit distance (with $95\%$ confidence intervals) vs. sample size on CHILD network.}
  \label{fg:child}
\end{figure}

\begin{figure}[h]
  \centering
    \includegraphics[scale=0.6]{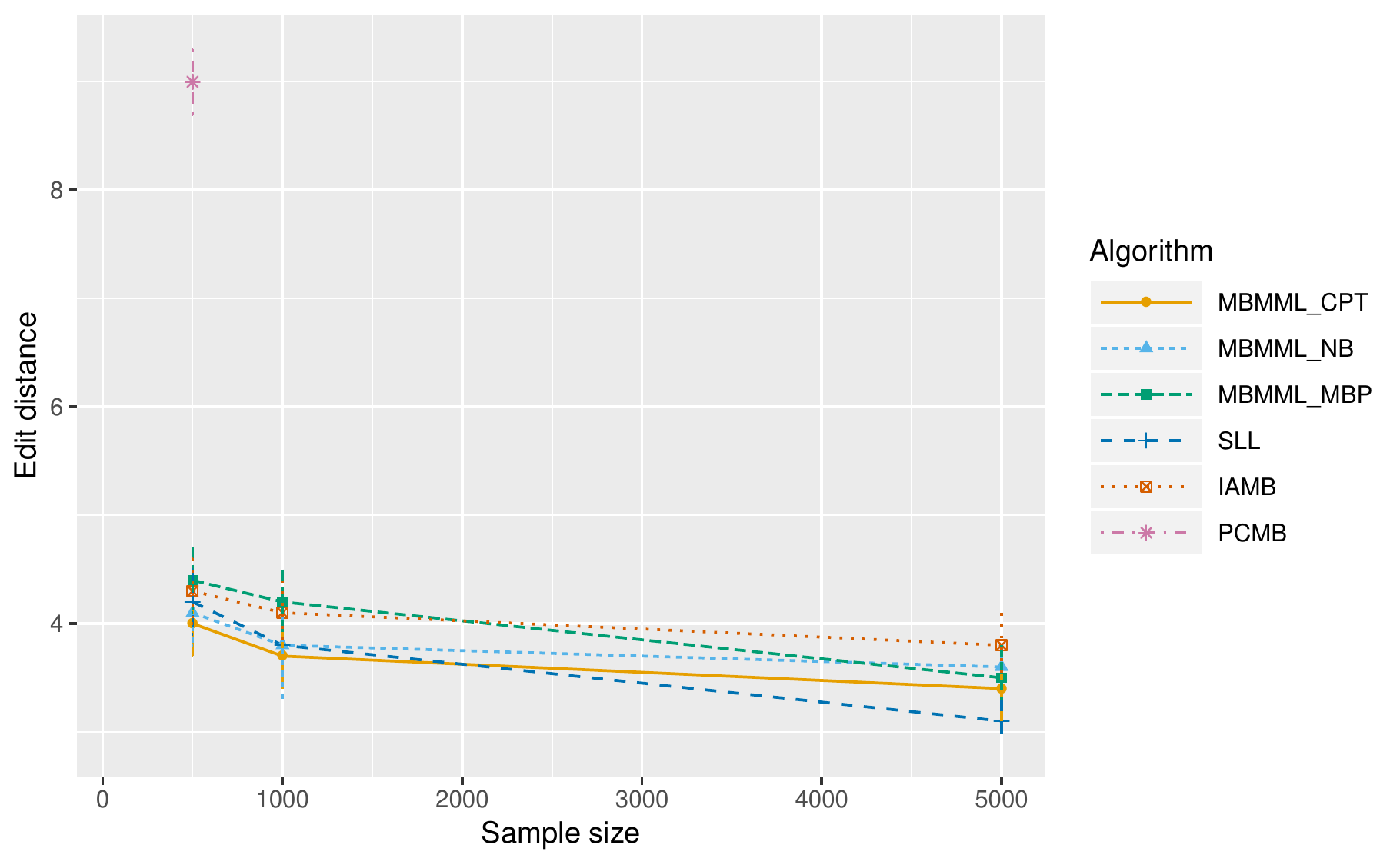}
  \caption{Edit distance (with $95\%$ confidence intervals) vs. sample size on BARLEY network. PCMB failed under 1000 and 5000 samples possibly due to an implementation error.}
  \label{fg:barley}
\end{figure}

Figures \ref{fg:child} and \ref{fg:barley} report the mean edit
distance (with confidence intervals) of all algorithms on the CHILD
and BARLEY networks.
\footnote{Note that PCMB failed to run on the BARLEY network for 1000 and 5000 sample sizes.}
In most cases, the algorithms show no statistically significant difference from each other, except that PCMB is less robust under small samples. Both PCMB
and SLL appear to converge slightly faster than the MML methods and
IAMB. The edit distance, precision and reacall on all six models
are summarized in Tables \ref{tb:all_ed} and \ref{tb:all_pre_rec} in
the Appendix. Although MBMML+MBP does not show competitive precision, but it has the highest recall in all cases with clear margins, especially under small samples. On these real networks SLL tends to outperform the alternative techniques, with MML+CPT being its equal or near equal.

%\textcolor{red}{an interesting finding is that for mml mbpt, mb ed reduced when taking intersection on small samples. so it kind of suggesting that when using mml mbpt, better to do intersection for small sample learning cases and union for large sample learning case. this behaviour is particularly obvious on real models, but not so obvious on artificial models.}

%------------------------------------------------------------------------------------------------------

\subsection{Accuracy on artificial models} 
\label{sec:exp_artificial}

It is also useful to test machine learning algorithms on artificial
models using generated data, where the ground truth is known.  The
DAGs we used tended to be more complex than the real models above. While
having similar numbers of variables, the fan-in and arity of variables
were somewhat higher (cf.~Table \ref{tab:exp_models}). Their
parameters were independently sampled from a uniform distribution
which matched the parameter prior used in the multi-state MML
metric. This could provide an advantage to the MML methods, since they
assume as much. However, as we discuss more below, we also tested the
MML methods with exactly the true prior for these models and found
very similar performance with them using the uniform prior assumption,
suggesting that MML performance doesn't vary much when the parameter
assumptions are approximately correct.

\begin{figure}[h]
  \centering
    \includegraphics[scale=0.6]{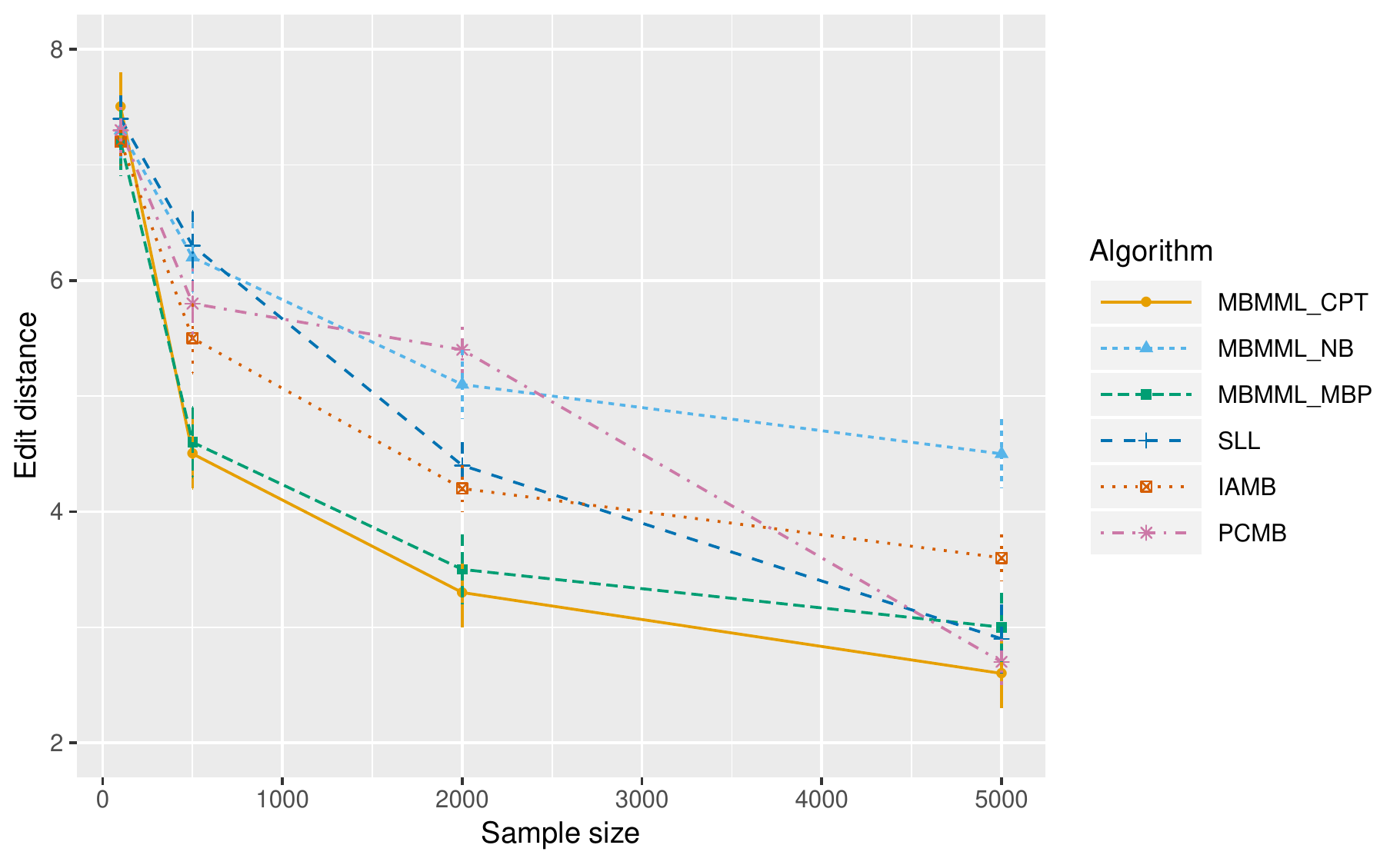}
  \caption{Edit distance (with $95\%$ confidence intervals) vs. sample size on artificial Bayesian networks (30-5-4-1) containing 30 variables, maximum 5 parents and maximum 4 states for each variable.}
  \label{fg:30}
\end{figure}\begin{figure}[H]
  \centering
    \includegraphics[scale=0.6]{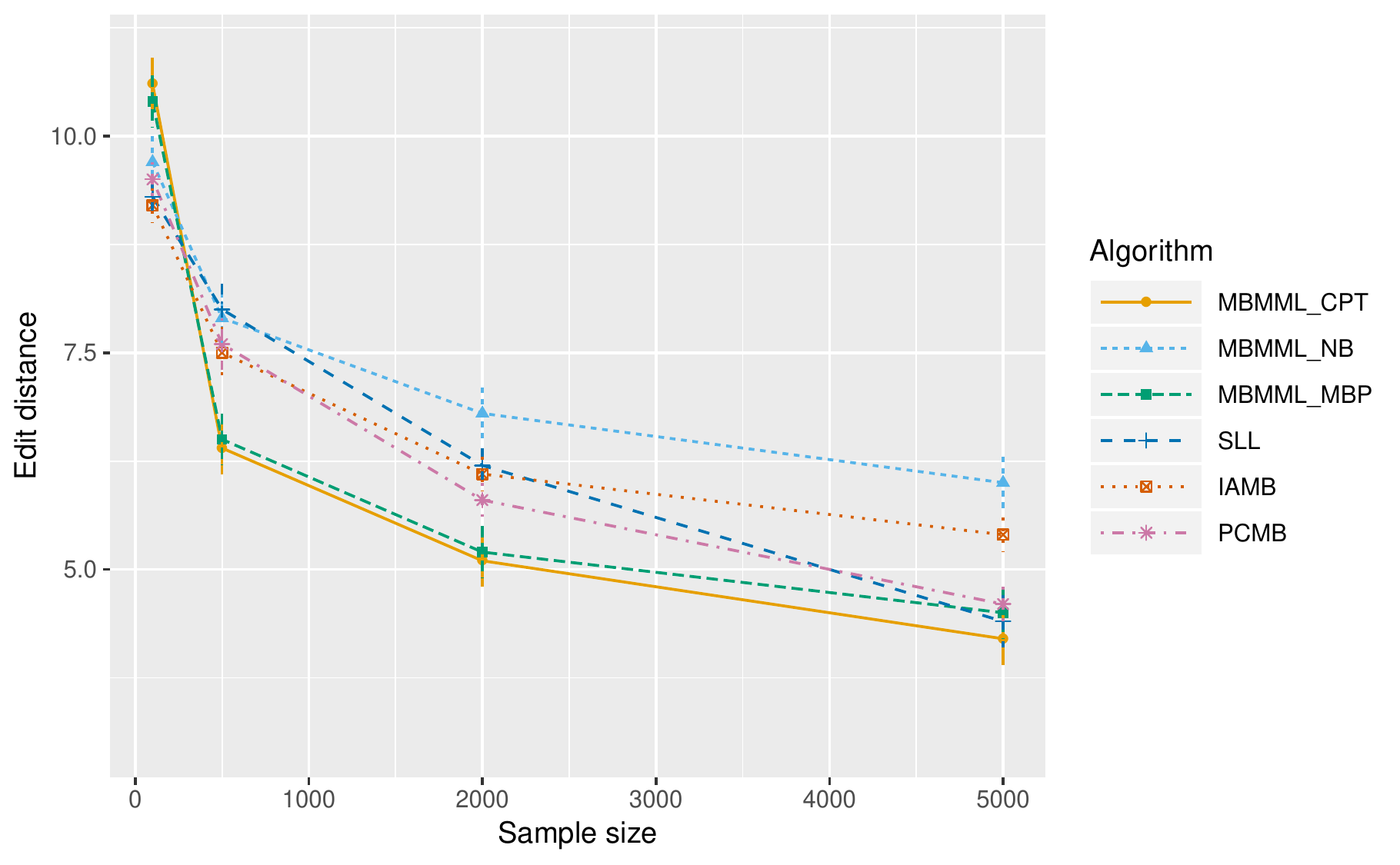}
  \caption{Edit distance (with $95\%$ confidence intervals) vs. sample size on artificial Bayesian networks (50-5-4-1) containing 50 variables, maximum 5 parents and maximum 4 states for each variable.}
  \label{fg:50}
\end{figure}

\begin{figure}[hbt]
  \centering
    \includegraphics[scale=0.6]{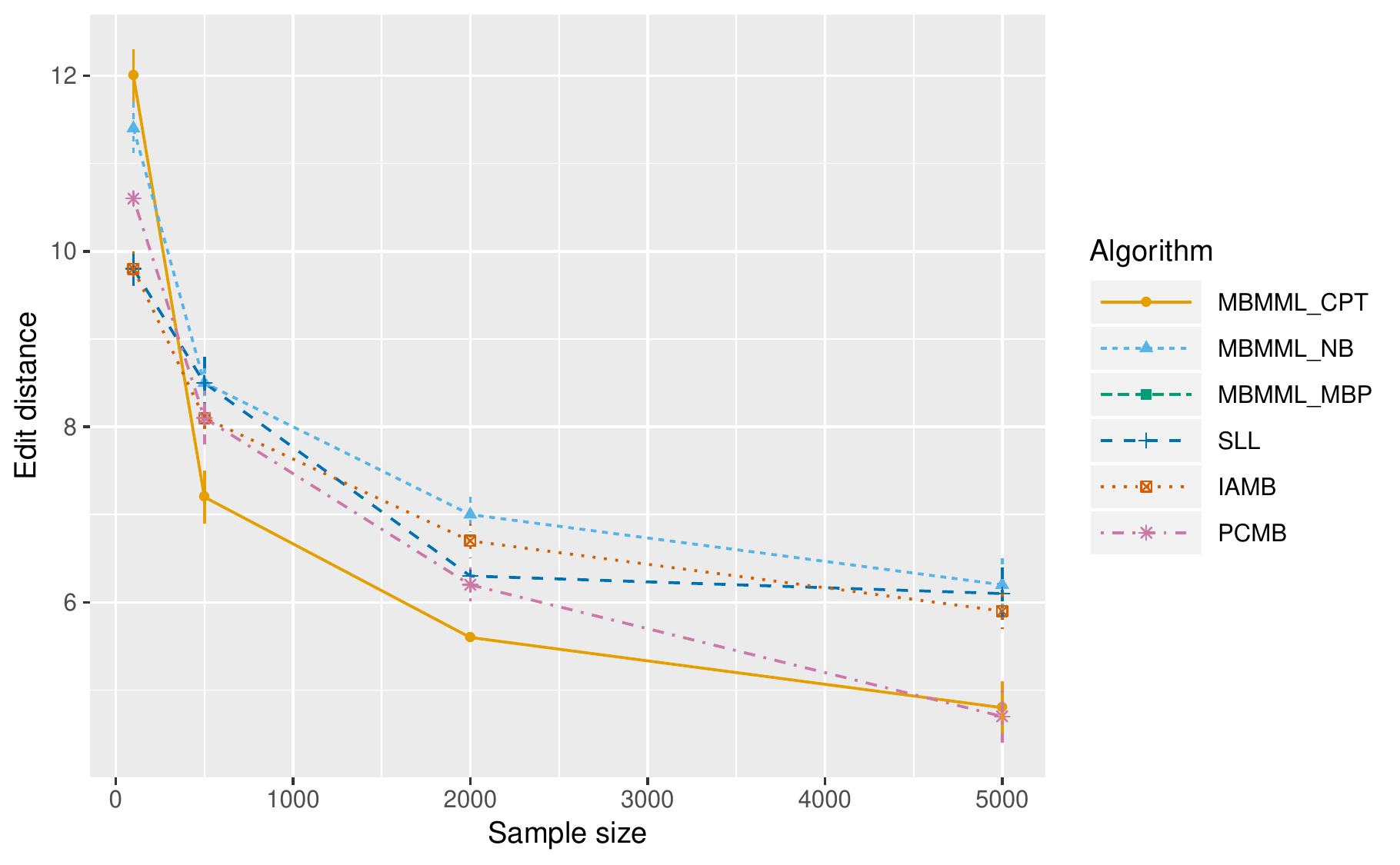}
  \caption{Edit distance (with $95\%$ confidence intervals) vs. sample size on artificial Bayesian networks (80-5-4-1) containing 80 variables, maximum 5 parents and maximum 4 states for each variable.}
  \label{fg:80}
\end{figure}

Figure \ref{fg:30} shows how the different algorithms perform with different sample sizes from the artificial networks. All algorithms perform similarly with very small samples (i.e., 100), while MBMML+NB and IAMB fall away from the pack at large samples (i.e., 5000). Ignoring MBMML+NB and IAMB, there is nothing to choose between the algorithms in terms of edit distance at 5000 samples. For moderate sample sizes (i.e., 500 and 2000),  MBMML+CPT and MBMML+MBP show significantly lower edit distances than the others. This suggests that the explanatory power of the CPT model is significantly improved when increasing the sample size from small to medium. Looking at precision and recall (see Appendix), MBMML+CPT and MBMML+MBP  have the highest recall in all cases, with MBMML+CPT's precision being the highest under medium and large samples. Figures \ref{fg:50} and \ref{fg:80} show similar trends as Figure \ref{fg:30}, except for the case of 100 samples where MBMML+CPT and MBMML+MBP appear to have higher edit distance than the others. Looking at their precision and recall (Table \ref{tb:all_pre_rec}) suggests that both models tend to overfit with 100 samples, but the problem is fixed by feeding in more data. Note that we did not run MBMML+MBP on 80-5-4-1 due to its high computational cost. In addition, SLL's edit distance does not drop much for 5000 samples, suggesting difficulties with larger 80-variable BNs even given large samples. It is worth pointing out that SLL did poorly for all 25 data sets of 5000 samples in this case. Possibly this is due to the program's large running time or space requirements, although the program falls back to an approximate algorithm for large MBs.
 
\begin{figure}[hbt]
  \centering
    \includegraphics[scale=0.6]{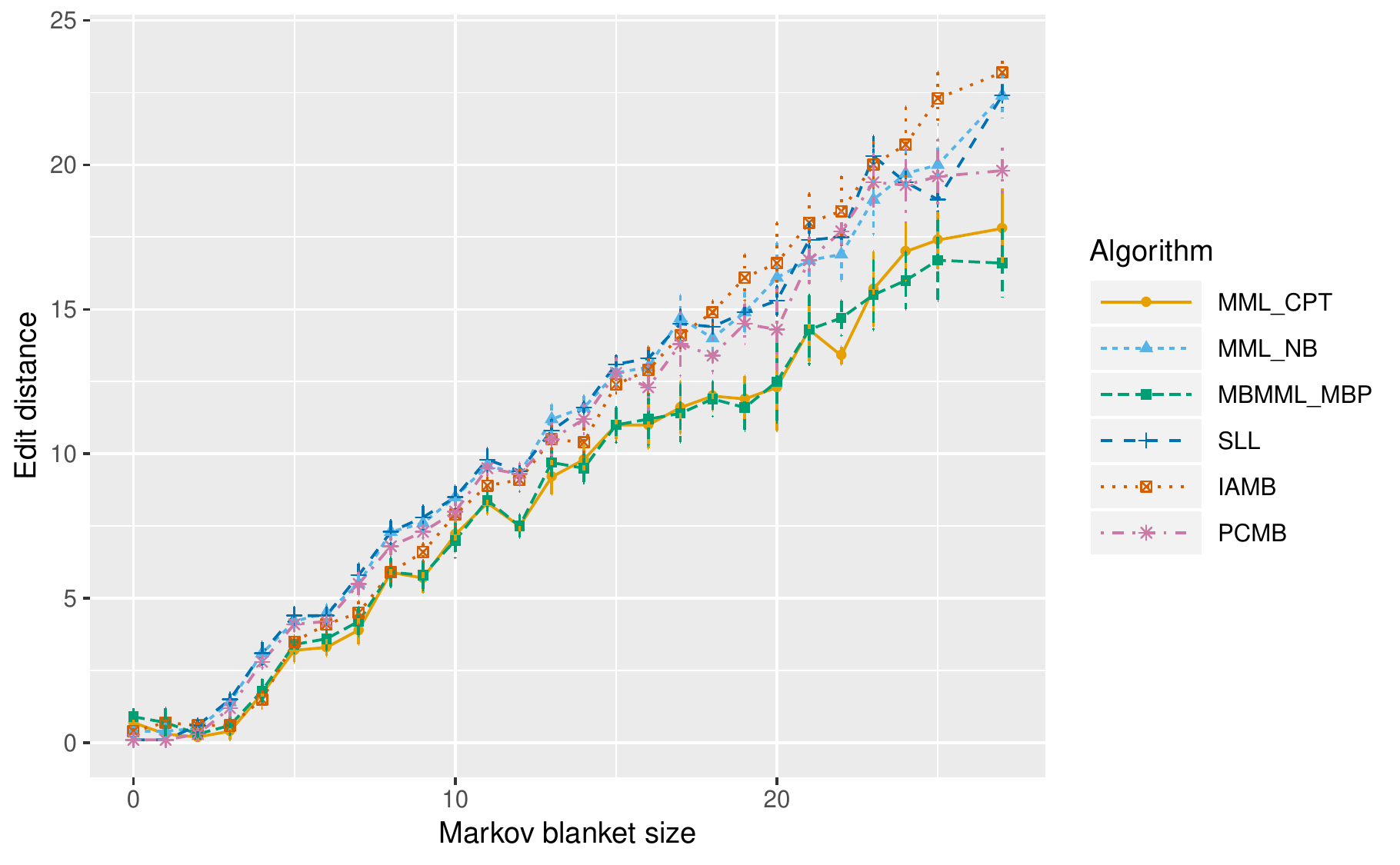}
  \caption{Edit distance versus average Markov Blanket size on 50-5-4-1 models with 500 samples.}
  \label{fg:ed_mb_50_500}
\end{figure} 

Figure \ref{fg:ed_mb_50_500} shows results for 50 variable networks
and a fixed sample size of 500. Average MB sizes are shown on the
X-axis; these were not controlled for, so the results simply reflect a
correlation between larger Markov Blankets and poorer accuracy, which
is to be expected of course. While on networks with small MBs all the
algorithms perform similarly, the MBMML+CPT and MBMML+MBP algorithms
are clearly outperforming the rest when MBs contain 20 or more
variables.\footnote{It is worth noting that for these larger MBs, SLL
  reverts to the GES algorithm.} That is, these MML algorithms clearly
recommend themselves for dealing with more complex discovery problems
with moderate sample sizes. Overall, MBMML+CPT and MBMML+MBP have the
lowest edit distances, which is consistent with the ranking in Figure
\ref{fg:50} for 500 samples.

\begin{figure}[hbt]
  \centering
    \includegraphics[scale=0.6]{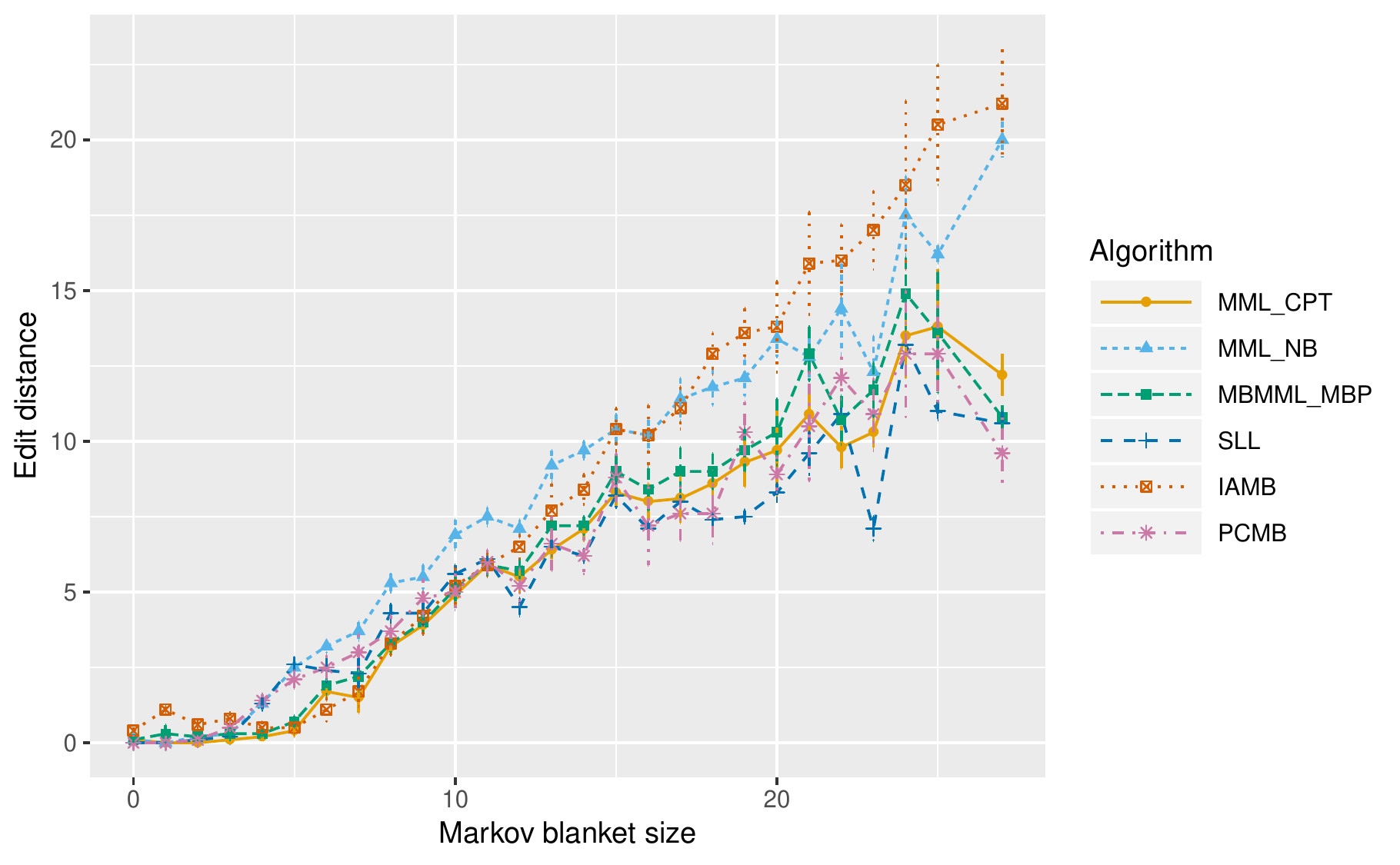}
  \caption{Edit distance versus Markov Blanket size on 50-5-4-1 models with 5000 samples.}
  \label{fg:ed_mb_50_5000}
\end{figure}

Upping the sample size to 5000 (Figure \ref{fg:ed_mb_50_5000}), we
find a somewhat different story. While MBMML+CPT/MBP are always
competitive, only occasionally being significantly worse than the best
performer per MB size (as shown by non-overlapping CIs), PCMB and SLL
(or GES) also are performing well across the board with the larger
samples. IAMB and MBMML+NB are pretty clearly underperforming at
larger samples and larger MBs, while MBMML+NB does well with smaller
MBs and large samples.

To sum up, IAMB has clearly been superceded by subsequently developed algorithms.
PCMB and SLL show some weaknesses in small and moderate sized samples,
but perform as well or better than alternatives given larger
samples. MBMML+NB does well with small samples, while MBMML+CPT/MBP
perform well across the board.

\subsubsection{Dirichlet priors}
\label{sec:exp_priors}

There may be a concern that if the generating models had non-uniform
parameter priors, the MML methods would perform differently. We
hypothesized that if the true priors are not uniform, then using
uniform priors would give noticeably worse results than using the true
prior. To be sure, performance will also depend on the quality and
size of the samples. To check the impact of using an uninformative,
uniform prior, MBMML+CPT was given both the true priors and uniform
priors when tested on a 30-5-4-1 network whose parameters were sampled
from a symmetric Dirichlet distribution with a non-uniform
concentration parameter
$\alpha \in \{0.1, 0.4, 0.7, 1, 10, 40, 70, 100\}$. The experiments
were done for 500 and 5000 samples.

\begin{figure}[hbt]
  \centering
    \includegraphics[scale=0.6]{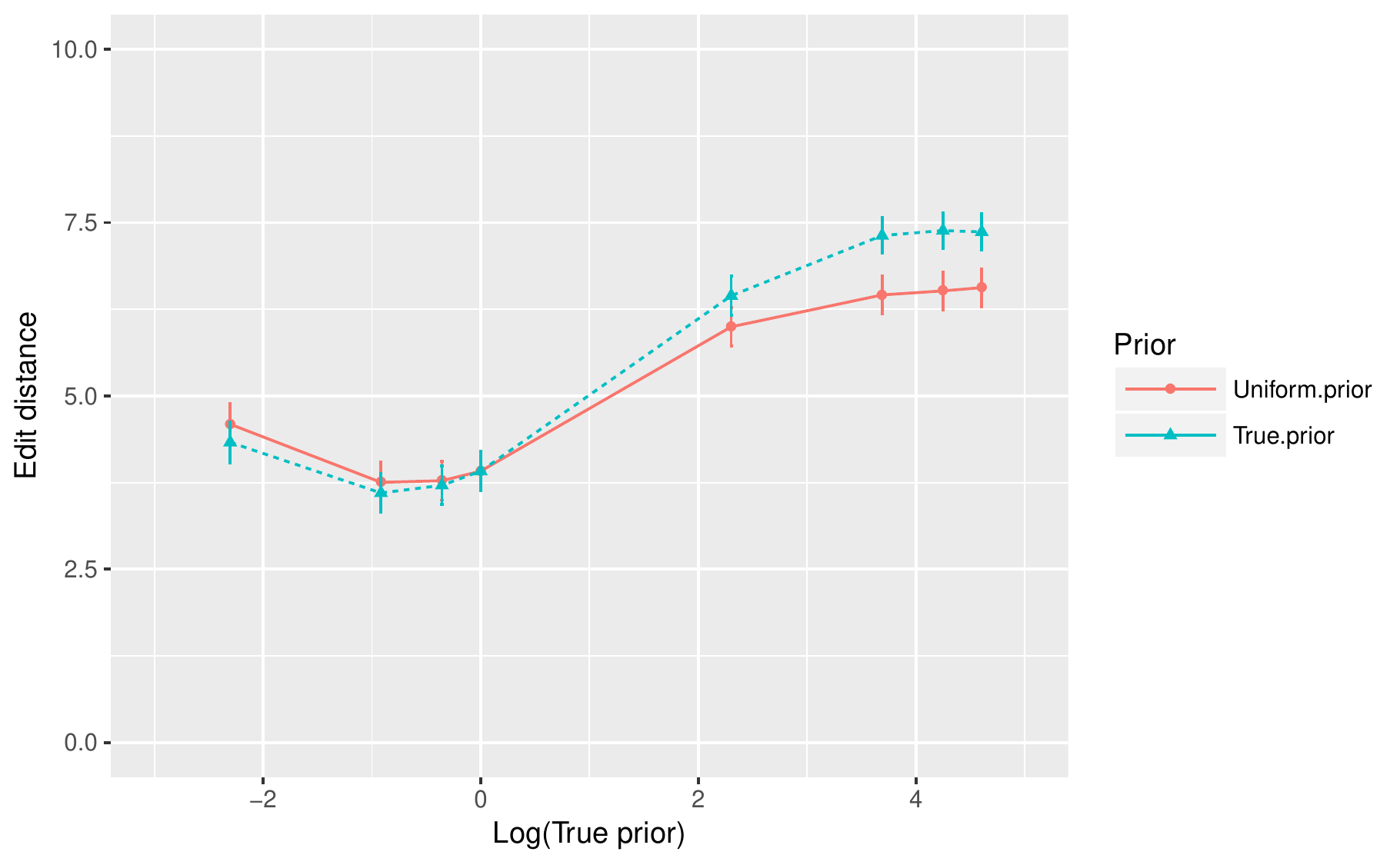}
  \caption{MBMML+CPT's edit distances using the true prior and uniform prior on a 30-5-4-1 model with 500 samples. The X-axis is the natural log scale of the true symmetric Dirichlet concentration parameter $\alpha = \{0.1, 0.4, 0.7, 1, 10, 40, 70, 100\}$.}
  \label{fg:wrong_prior_500}
\end{figure}

\begin{figure}[hbt]
  \centering
    \includegraphics[scale=0.6]{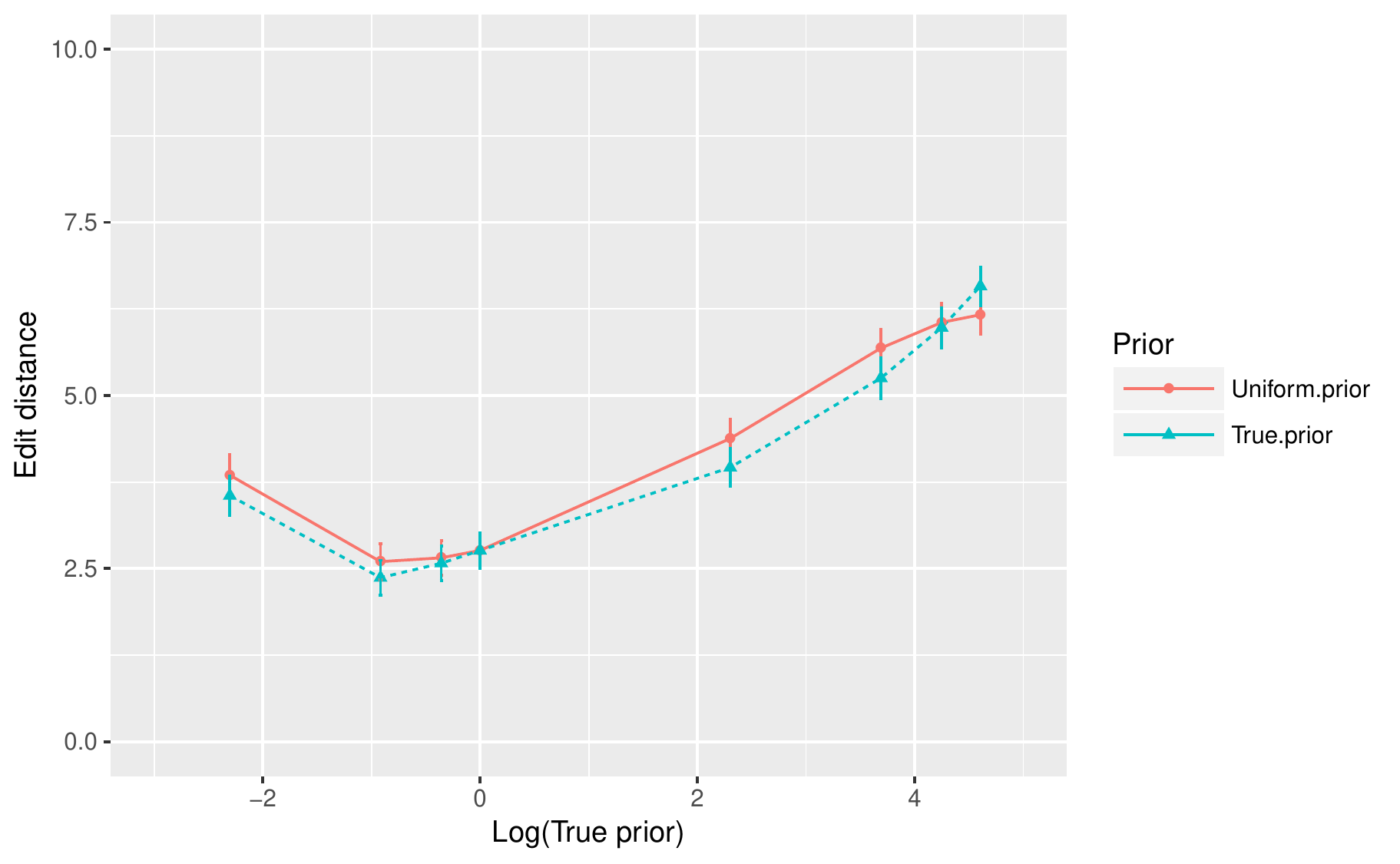}
  \caption{MBMML+CPT's edit distances using the true prior and uniform prior on a 30-5-4-1 model with 5000 samples. The X-axis is the natural log scale of the true symmetric Dirichlet concentration parameter $\alpha = \{0.1, 0.4, 0.7, 1, 10, 40, 70, 100\}$.}
  \label{fg:wrong_prior_5000}
\end{figure}

Figures \ref{fg:wrong_prior_500} and \ref{fg:wrong_prior_5000} report
no significant differences between the use of true priors and uniform
priors when the concentration parameters $\alpha \le 1$. This is
because adding small $\alpha$ values to parameter estimates will have
a small effect, swamped even by modest data. When $\alpha > 1$,
uniform priors track the true priors until the non-uniformity begins
to get extreme ($\alpha \ge 10$, i.e., $\ln \alpha > 2$).

As $\alpha$ increases, the learned Markov Blankets increase in
complexity. This is analogous to the increased complexity of models
learned using the BDeu metric when increasing its equivalent sample size
\citep{silander2007sensitivity}. The MML metric for the CPT model with
symmetric Dirichlet priors is similar to the BDeu metric, although MML
includes costs for the precision of parameter estimates. But both
metrics penalize model complexity using a function of $\alpha$, which
decreases as $\alpha$ increases. Hence, given larger $\alpha$, MML
methods more easily discover larger Markov Blankets. These may contain
a larger proportion of false positives, especially with small
samples. At larger samples, these differences between MML with uniform
and true priors, however, appear to be erased.

In general, we didn't find important differences between MML-CPT with
uniform and with true priors, supporting our use of the practical, and
non-informative, uniform priors.

\begin{table}[htb]
\centering
\caption{MB discovery algorithm complexity}
\label{tb:bigo}
\begin{tabular}{ll}
\hline
Algorithm    & Big O notation \\ \hline
IAMB      & $O(n^2)$      \\
MBMML+NB          &    $O(n^2)$    \\
MBMML+CPT          &   $O(n2^{n-1})$      \\
MBMML+MBP         & $O(n2^{n-1})$        \\
PCMB & $O(n^2 2^{n-1})$     \\
SLL        & $O(n^4 2^n)$       \\ \hline 
\end{tabular}
\end{table}

\subsection{Algorithmic complexity}
Table \ref{tb:bigo} orders all algorithms by ascending computational
complexity. The main loop in Algorithm \ref{alg:mbmmlf} for
MBMML+CPT/NB runs at most $n-1$ times. Each time it runs through all
unchecked nodes to find the best candidate to add to the MB using the
MML metric. For a CPT model, there can be at most $n-1$ parents, with
the multi-state MML summed over all $2^{n-1}$ parent
instantiations. So the computational complexity of the MBMML+CPT
algorithm is $O(n2^{n-1})$. For an NB model, the worst case is when
all $n-1$ nodes are children of the target, which is linear in $n$
within the \textit{WHILE} loop, and so gives a complexity of
$O(n^2)$. The worst case for MBMML+MBP is when one of the sampled
polytrees is just the full CPT model, with all MB variables parents of
the target. In general, a random polytree model is slower than a CPT
by a constant factor, which is determined by the number of sampled
regional structures. But this doesn't affect the O-notation
complexity.

For PCMB, the total time required is dominated by the process of
finding the direct neighbors of the target. This process tries to find
a subset of the neighbor set, conditioning on which the target is
independent with a candidate. And such a process runs through all
variables to ensure the symmetry property holds. Hence, its complexity in
the worst case is $O(n^2 2^{n-1})$. The total time required by IAMB
and SLL were published in the associated papers, by
\citet{tsamardinos2003algorithms} and \citet{niinimaki2012local}
respectively.\footnote{\citet{gao2017efficient} made significant
  practical gains in SLL's performance by relaxing symmetry
  enforcement, but we did not explore that here.}
  
%------------------------------------------------------------------------------------------------------
%------------------------------------------------------------------------------------------------------

\section{Conclusion}
\label{sec:disc}

We have proposed and tested three alternative MML methods for learning
Markov Blankets. The three methods all use the multi-state MML
measure, but apply them with different models, namely a CPT model,
Naive Bayes and an ensemble of random Markov Blanket polytrees. We
proved that the MBMML+CPT algorithm will find the correct Markov
Blankets given perfect data (i.e., infinite samples), although it will
not be data efficient for large Markov Blankets due to the exponential
number of parameters required. We looked at one of the more common
answers to data inefficiency in Naive Bayes models, which sacrifice
the modeling of conditional dependencies for speed and simplicity. As
a compromise between the correctness but inefficiency of CPTs and the
efficiency but strong assumptions of NB, we also explored an ensemble
technique in MBMML+MBP, using random polytrees within Markov Blankets.

We tested these three MML algorithms against three of the best
alternatives reported in recent literature, with both real data and
artificial Bayesian networks at a range of sample sizes. Our empirical
results show that overall neither MBMML+NB nor IAMB are competitive in
terms of edit distance with the alternative algorithms at medium to
large sample sizes. MBMML+CPT/MBP, PCMB, and SLL are competitive with
each other at large samples, with some advantage shown by the MML
methods at smaller samples and larger model complexities. In terms of
time complexity, MML has the best worst case performance of these
algorithms. The MML methods appear to be worth applying in practice
across a range of problems.

\newpage

%------------------------------------------------------------------------------------------------------
%------------------------------------------------------------------------------------------------------

\section{Appendix}
\label{sec:Appendix}

%\begin{landscape}
\begin{table}[]
\centering
\caption{Summary of edit distance (with $95\%$ confidence intervals) of all Markov Blanket discovery algorithms on both real and artificial Bayesian networks. The best results are highlighted in pink. In real networks, SLL wins in all cases followed by MBMML+CPT, PCMB, MBMML+NB, IAMB and MMLL+MBP. PCMB failed to learn on BARLEY networks with 1000 and 5000 samples possibly due to an implementation error. In artificial networks, MBMML+CPT and MBMML+MBP win most of the times followed by SLL/PCMB and MBMML+NB/IAMB. We did not run MBMML+MBP on 80-5-4-1 due to its high computation cost. Note that SLL's edit distance is much worse on 80-5-4-1 under 5000 samples. This is due to its inability of learning local structures on BNs of this size with 5000 samples.}
\label{tb:all_ed}
\scalebox{0.8}{
\begin{tabular}{llllllll}
\hline
Network    & SAMPLES &  \begin{tabular}[c]{@{}l@{}}$MBMML$\\ $+CPT$\end{tabular} &  \begin{tabular}[c]{@{}l@{}}$MBMML$\\ $+NB$\end{tabular} & \begin{tabular}[c]{@{}l@{}}$MBMML$\\ $+MBP$\end{tabular} & IAMB     & PCMB     & SLL      \\ \hline
CHILD      & 500     & \cellcolor{babypink}0.9+-0.2      & \cellcolor{babypink}0.9+-0.2     & \cellcolor{babypink}1.3+-0.2  & \cellcolor{babypink}1.2+-0.2 & \cellcolor{babypink}1.3+-0.2 & \cellcolor{babypink}1+-0.2   \\
           & 1000    & \cellcolor{babypink}0.7+-0.1      & \cellcolor{babypink}0.7+-0.2     & \cellcolor{babypink}0.9+-0.1  & 1.1+-0.2 & \cellcolor{babypink}1+-0.2   & \cellcolor{babypink}0.8+-0.1 \\
           & 5000    & 0.5+-0.1      & 0.6+-0.1     & \cellcolor{babypink}0.2+-0.1  & 0.7+-0.2 & \cellcolor{babypink}0.1+-0 & \cellcolor{babypink}0.2+-0.1 \\ \hline
INSURANCE  & 500     & \cellcolor{babypink}3.3+-0.2      & \cellcolor{babypink}3.5+-0.2     & 4+-0.3  & \cellcolor{babypink}3.4+-0.3 & \cellcolor{babypink}3.2+-0.2 & \cellcolor{babypink}3.1+-0.2 \\
           & 1000    & \cellcolor{babypink}2.9+-0.2      & 3.3+-0.2     & 3.5+-0.3  & \cellcolor{babypink}3.1+-0.3 & \cellcolor{babypink}2.9+-0.2   & \cellcolor{babypink}2.7+-0.2 \\
           & 5000    & \cellcolor{babypink}2.1+-0.2      & 2.8+-0.2     & 2.4+-0.2  & 2.7+-0.2 & \cellcolor{babypink}1.8+-0.2 & \cellcolor{babypink}2+-0.2   \\ \hline
ALARM      & 500     & 1.4+-0.1      & 2.1+-0.2     & 3.4+-0.2  & 1.9+-0.2 & 1.5+-0.1 & \cellcolor{babypink}0.8+-0.1 \\
           & 1000    & 1+-0.1        & 1.8+-0.2     & 2.8+-0.2  & 1.6+-0.2   & 1.1+-0.1 & \cellcolor{babypink}0.6+-0.1 \\
           & 5000    & 0.5+-0.1      & 1.5+-0.2     & 1.7+-0.1  & 1.3+-0.2 & \cellcolor{babypink}0.3+-0.1 & \cellcolor{babypink}0.2+-0   \\ \hline
BARLEY     & 500     & \cellcolor{babypink}4+-0.3        & \cellcolor{babypink}4.1+-0.3     & \cellcolor{babypink}4.4+-0.3  & \cellcolor{babypink}4.3+-0.3 & 9+-0.5 & \cellcolor{babypink}4.2+-0.2 \\
           & 1000    & \cellcolor{babypink}3.7+-0.3      & \cellcolor{babypink}3.8+-0.3     & \cellcolor{babypink}4.2+-0.3  & \cellcolor{babypink}4.1+-0.3 & NA & \cellcolor{babypink}3.8+-0.2 \\
           & 5000    & \cellcolor{babypink}3.4+-0.3      & \cellcolor{babypink}3.6+-0.3     & \cellcolor{babypink}3.5+-0.3  & 3.8+-0.3 & NA & \cellcolor{babypink}3.1+-0.2 \\ \hline
HAILFINDER & 500     & \cellcolor{babypink}4.4+-0.3      & \cellcolor{babypink}4.3+-0.2      & 5.2+-0.3  & \cellcolor{babypink}4.1+-0.2 & 7.1+-0.5 & \cellcolor{babypink}4.3+-0.3 \\
           & 1000    & \cellcolor{babypink}4.4+-0.3      & \cellcolor{babypink}4.3+-0.2     & 5+-0.3    & \cellcolor{babypink}4.1+-0.2 & 6.2+-0.4 & \cellcolor{babypink}4.1+-0.3 \\
           & 5000    & \cellcolor{babypink}4.3+-0.3      & \cellcolor{babypink}4.3+-0.2       & 5.1+-0.3  & \cellcolor{babypink}4.2+-0.2 & \cellcolor{babypink}3.8+-0.2 & \cellcolor{babypink}4+-0.3  \\ \hline
30-5-4-1   & 100     & \cellcolor{babypink}7.5+-0.3      & \cellcolor{babypink}7.3+-0.3     & \cellcolor{babypink}7.2+-0.3  & \cellcolor{babypink}7.2+-0.3 & \cellcolor{babypink}7.3+-0.3 & \cellcolor{babypink}7.4+-0.3 \\
		   & 500     & \cellcolor{babypink}4.5+-0.3      & 6.2+-0.3     & \cellcolor{babypink}4.6+-0.3  & 5.5+-0.3 & 5.8+-0.3 & 6.3+-0.3 \\
           & 2000    & \cellcolor{babypink}3.3+-0.2      & 5.1+-0.3     & \cellcolor{babypink}3.5+-0.2  & 4.2+-0.3 & 5.4+-0.2 & 4.4+-0.3 \\
           & 5000    & \cellcolor{babypink}2.6+-0.2      & 4.5+-0.2     & \cellcolor{babypink}3+-0.2    & 3.6+-0.3 & \cellcolor{babypink}2.7+-0.2 & \cellcolor{babypink}2.9+-0.2 \\ \hline
50-5-4-1   & 100     & 10.66+-0.3      & \cellcolor{babypink}9.7+-0.3     & 10.4+-0.3  & \cellcolor{babypink}9.2+-0.3 & \cellcolor{babypink}9.5+-0.3 & \cellcolor{babypink}9.3+-0.3   \\
		   & 500     & \cellcolor{babypink}6.4+-0.3      & 7.9+-0.3     & \cellcolor{babypink}6.5+-0.3  & 7.5+-0.3 & 7.6+-0.3 & 8+-0.3 \\
           & 2000    & \cellcolor{babypink}5.1+-0.2      & 6.8+-0.3     & \cellcolor{babypink}5.2+-0.2  & 6.1+-0.3 & 5.8+-0.2 & 6.2+-0.3 \\
           & 5000    & \cellcolor{babypink}4.2+-0.2      & 6+-0.2     & \cellcolor{babypink}4.5+-0.2  & 5.4+-0.3 & \cellcolor{babypink}4.6+-0.2 & \cellcolor{babypink}4.4+-0.2 \\ \hline
80-5-4-1   & 100     & 12+-0.3      & 11.4+-0.3  & NA   & \cellcolor{babypink}9.8+-0.3  & 10.6+-0.3 & \cellcolor{babypink}9.8+-0.3 \\
           & 500     & \cellcolor{babypink}7.2+-0.2      & 8.5+-0.3     & NA  & 8.1+-0.3 & 8.1+-0.3 & 8.5+-0.3   \\
           & 2000    & \cellcolor{babypink}5.6+-0.2      & 7+-0.3     & NA  
           & 6.7+-0.3 & 6.2+-0.2 & 6.3+-0.3 \\
           & 5000    & \cellcolor{babypink}4.8+-0.2      & 6.2+-0.2     & NA  
           & 5.9+-0.3 & \cellcolor{babypink}4.7+-0.2 & 6.1+-0.3 \\ \hline           
\end{tabular}
}
\end{table}
%\end{landscape}

\begin{landscape}
\begin{table}[]
\begin{center}
\caption{Summary of precision and recall (with $95\%$ confidence intervals) of all Markov Blanket discovery algorithms on both real and artificial Bayesian networks. The best precisions and recalls are respecitively highlighted in pink and blue. In real networks, IAMB almost always has the highest precision, followed by SLL, MBMML+CPT/NB, PCMB and MBMML+MBP. However, MBMML+CPT has the highest recall across all cases by significant margins. In artificial networks, MBMML+CPT has the highest precision in all medium and large sample cases, followed by IAMB who has the highest precision in small samples. In terms of recall, MBMML+CPT and MBMML+MBP both win in all cases. We did not run MBMML+MBP on 80-5-4-1 due to its high computation cost. Note that SLL's edit distance is much worse on 80-5-4-1 under 5000 samples. This is due to its inability of learning local structures on BNs of this size with 5000 samples.}
\label{tb:all_pre_rec}
\scalebox{0.7}{
\begin{tabular}{llllllllllllll}
\hline
Network    & SAMPLES & \multicolumn{2}{c}{MBMML+CPT}                          & \multicolumn{2}{c}{MBMML+NB}                           & \multicolumn{2}{c}{MBMML+MBP}                              & \multicolumn{2}{c}{IAMB}                                   & \multicolumn{2}{c}{PCMB}                                   & \multicolumn{2}{c}{SLL}                                    \\
           &         & \multicolumn{1}{c}{Precision} & \multicolumn{1}{c}{Recall} & \multicolumn{1}{c}{Precision} & \multicolumn{1}{c}{Recall} & \multicolumn{1}{c}{Precision} & \multicolumn{1}{c}{Recall} & \multicolumn{1}{c}{Precision} & \multicolumn{1}{c}{Recall} & \multicolumn{1}{c}{Precision} & \multicolumn{1}{c}{Recall} & \multicolumn{1}{c}{Precision} & \multicolumn{1}{c}{Recall} \\ \hline
CHILD      & 500     & \cellcolor{babypink}0.94+-0.03                    & 0.8+-0.04                  & \cellcolor{babypink}0.94+-0.03                    & \cellcolor{cyan}0.82+-0.04                 & 0.78+-0.04                    & \cellcolor{cyan}0.89+-0.03                 & \cellcolor{babypink}0.95+-0.03                    & 0.77+-0.04                 & 0.87+-0.04                    & 0.77+-0.04                 & \cellcolor{babypink}0.94+-0.03                    & 0.78+-0.04                 \\
           & 1000    
           & \cellcolor{babypink}0.98+-0.02                    & 0.88+-0.03                 
           & \cellcolor{babypink}0.97+-0.02                    & 0.86+-0.03                 
           & 0.82+-0.03                    & \cellcolor{cyan}0.96+-0.01                 
           & \cellcolor{babypink}0.95+-0.02                    & 0.83+-0.03                 
           & \cellcolor{babypink}0.94+-0.03                    & 0.8+-0.04                 
           & \cellcolor{babypink}0.97+-0.02                    & 0.84+-0.03                 \\
           & 5000    
           & \cellcolor{babypink}1+-0.01                       & 0.91+-0.02                 
           & \cellcolor{babypink}1+-0.01                       & 0.89+-0.02                 
           & 0.95+-0.02                    & \cellcolor{cyan}1+-0                       
           & \cellcolor{babypink}0.98+-0.01                    & 0.9+-0.02                 
           & \cellcolor{babypink}1+-0                    & \cellcolor{cyan}0.99+-0.01                 
           & \cellcolor{babypink}1+-0                          & 0.97+-0.01                 \\ \hline
INSURANCE  & 500     
& 0.82+-0.04                    & 0.48+-0.03                
& 0.8+-0.04                     & 0.42+-0.03                 
& 0.64+-0.04                     & \cellcolor{cyan}0.67+-0.03                  
& \cellcolor{babypink}0.9+-0.03                    & 0.43+-0.03                 
& 0.77+-0.04                    & 0.5+-0.03                 
& \cellcolor{babypink}0.83+-0.04                    & 0.51+-0.04                 \\
           & 1000    
           & 0.86+-0.03                    & 0.54+-0.03                 
           & 0.85+-0.04                    & 0.45+-0.03                 
           & 0.65+-0.03                    & \cellcolor{cyan}0.73+-0.03                
           & \cellcolor{babypink}0.93+-0.03                    & 0.5+-0.03                  
           & 0.81+-0.04                    & 0.53+-0.03                
           & \cellcolor{babypink}0.88+-0.03                    & 0.58+-0.03                 \\
           & 5000   
           & 0.95+-0.01                    & 0.68+-0.03                 
           & 0.93+-0.02                    & 0.57+-0.03                 
           & 0.78+-0.03                    & \cellcolor{cyan}0.8+-0.03                 
           & \cellcolor{babypink}0.97+-0.02                    & 0.59+-0.03                
           & \cellcolor{babypink}0.94+-0.03                    & 0.68+-0.03                
           & \cellcolor{babypink}0.98+-0.01                    & 0.69+-0.03                 \\ \hline
ALARM      & 500    
& 0.85+-0.03                    & 0.77+-0.03                 
& 0.79+-0.03                    & 0.67+-0.03              
& 0.53+-0.03                    & \cellcolor{cyan}0.92+-0.02              
& \cellcolor{babypink}0.9+-0.02                    & 0.62+-0.03              
& 0.85+-0.03                    & 0.7+-0.03              
& \cellcolor{babypink}0.92+-0.02                    & \cellcolor{cyan}0.89+-0.02                 \\
           & 1000 
           & 0.9+-0.02                     & 0.82+-0.03            
           & 0.86+-0.02                    & 0.68+-0.03             
           & 0.59+-0.03                    & \cellcolor{cyan}0.94+-0.02             
           & \cellcolor{babypink}0.96+-0.01                    & 0.71+-0.03            
           & 0.9+-0.03                     & 0.79+-0.03             
           & \cellcolor{babypink}0.94+-0.01                    &\cellcolor{cyan} 0.94+-0.01                 \\
           & 5000   
           & 0.97+-0.01                    & 0.93+-0.02        
           & 0.95+-0.02                    & 0.7+-0.03            
           & 0.7+-0.02                    & \cellcolor{cyan}0.97+-0.01           
           & 0.94+-0.02                    & 0.79+-0.03           
           & \cellcolor{babypink}1+-0.01                          & 0.95+-0.01          
           & \cellcolor{babypink}0.98+-0.01                    & \cellcolor{cyan}0.98+-0.01                 \\ \hline
BARLEY     & 500    
& \cellcolor{babypink}0.74+-0.03                    & 0.37+-0.03           
& \cellcolor{babypink}0.74+-0.03                    & 0.35+-0.02               
& 0.66+-0.03                    & \cellcolor{cyan}0.51+-0.03               
& \cellcolor{babypink}0.79+-0.04                            & 0.29+-0.02               
& 0.32+-0.02                            & \cellcolor{cyan}0.56+-0.03              
& 0.63+-0.04                    & 0.25+-0.02                 \\
           & 1000    
           & \cellcolor{babypink}0.79+-0.03                    & 0.42+-0.03        
           & \cellcolor{babypink}0.79+-0.03                    & 0.37+-0.02           
           & 0.68+-0.03                    & \cellcolor{cyan}0.57+-0.03          
           & \cellcolor{babypink}0.8+-0.03                            & 0.33+-0.03        
           & NA                            & NA                       
           & 0.72+-0.04                    & 0.35+-0.03                 \\
           & 5000   
           & \cellcolor{babypink}0.8+-0.03                     & 0.52+-0.03              
           & \cellcolor{babypink}0.81+-0.03                    & 0.47+-0.02             
           & 0.72+-0.03                    & \cellcolor{cyan}0.7+-0.03             
           & \cellcolor{babypink}0.84+-0.03                            & 0.38+-0.03         
           & NA                            & NA                     
           & \cellcolor{babypink}0.85+-0.03                    & 0.5+-0.03                  \\ \hline
HAILFINDER & 500     
& \cellcolor{babypink}0.3+-0.03                     & \cellcolor{cyan}0.18+-0.02             
& 0.25+-0.03                    & 0.12+-0.01              
& \cellcolor{babypink}0.27+-0.03                    & \cellcolor{cyan}0.2+-0.02              
& \cellcolor{babypink}0.32+-0.03                            & \cellcolor{cyan}0.18+-0.02         
& \cellcolor{babypink}0.3+-0.03                            & \cellcolor{cyan}0.18+-0.02            
& \cellcolor{babypink}0.28+-0.03                    & 0.14+-0.02                 \\
           & 1000   
           & \cellcolor{babypink}0.31+-0.03                    & \cellcolor{cyan}0.22+-0.02      
           & 0.26+-0.03                    & 0.12+-0.01          
           & \cellcolor{babypink}0.29+-0.03                    & \cellcolor{cyan}0.24+-0.02           
           & \cellcolor{babypink}0.34+-0.03                            & \cellcolor{cyan}0.2+-0.02          
           & \cellcolor{babypink}0.34+-0.03                            & \cellcolor{cyan}0.2+-0.02            
           & \cellcolor{babypink}0.3+-0.03                     & 0.18+-0.02                 \\
           & 5000   
           & \cellcolor{babypink}0.34+-0.03                    & \cellcolor{cyan}0.26+-0.02          
           & 0.26+-0.03                    & 0.14+-0.02          
           & \cellcolor{babypink}0.3+-0.03                     & \cellcolor{cyan}0.27+-0.03           
           & \cellcolor{babypink}0.33+-0.02                            & \cellcolor{cyan}0.22+-0.02             
           & \cellcolor{babypink}0.32+-0.03                            & 0.21+-0.02             
           & \cellcolor{babypink}0.34+-0.03                    & \cellcolor{cyan}0.22+-0.02                 \\ \hline
30-5-4-1   & 100    
& 0.56+-0.02                    & \cellcolor{cyan}0.36+-0.02              
& \cellcolor{babypink}0.6+-0.03                     & 0.23+-0.02              
& 0.58+-0.02                    & \cellcolor{cyan}0.36+-0.02               
& \cellcolor{babypink}0.65+-0.03                    & 0.21+-0.02               
& \cellcolor{babypink}0.59+-0.03                    & 0.25+-0.02              
& 0.5+-0.03                     & 0.17+-0.02                 \\
           & 500    
           & \cellcolor{babypink}0.91+-0.02                    & \cellcolor{cyan}0.56+-0.02           
           & 0.86+-0.02                    & 0.35+-0.02            
           & 0.86+-0.02                    & \cellcolor{cyan}0.56+-0.02            
           & \cellcolor{babypink}0.9+-0.02                    & 0.48+-0.02              
           & \cellcolor{babypink}0.89+-0.02                    & 0.38+-0.02             
           & 0.79+-0.03                    & 0.3+-0.02                  \\
           & 2000   
           & \cellcolor{babypink}0.97+-0.01                    & \cellcolor{cyan}0.68+-0.02        
           & 0.94+-0.01                    & 0.48+-0.02        
           & 0.94+-0.01                    & \cellcolor{cyan}0.68+-0.02        
           & \cellcolor{babypink}0.95+-0.01                    & \cellcolor{cyan}0.64+-0.02         
           & 0.6+-0.03                    & 0.38+-0.03          
           & \cellcolor{babypink}0.94+-0.02                    & 0.54+-0.02                 \\
           & 5000   
           & \cellcolor{babypink}0.99+-0                       & \cellcolor{cyan}0.76+-0.02        
           & 0.96+-0.01                    & 0.57+-0.02          
           & 0.96+-0.01                    & \cellcolor{cyan}0.73+-0.02            
           & 0.96+-0.01                    & 0.71+-0.02           
           & 0.94+-0.01                    & \cellcolor{cyan}0.77+-0.02           
           & \cellcolor{babypink}0.98+-0.01                    & 0.7+-0.02                  \\ \hline
50-5-4-1   & 100    
& 0.44+-0.02                    & \cellcolor{cyan}0.28+-0.01           
& 0.47+-0.02                    & 0.19+-0.01             
& 0.42+-0.02                    & \cellcolor{cyan}0.27+-0.01               
& \cellcolor{babypink}0.56+-0.03                    & 0.17+-0.01                
& 0.48+-0.02                    & 0.18+-0.01               
& 0.45+-0.03                    & 0.12+-0.01                 \\
           & 500   
           & \cellcolor{babypink}0.85+-0.02                    & \cellcolor{cyan}0.46+-0.02               
           & 0.77+-0.02                    & 0.29+-0.02           
           & 0.8+-0.02                     & \cellcolor{cyan}0.46+-0.02           
           & \cellcolor{babypink}0.83+-0.02                    & 0.38+-0.02           
           & \cellcolor{babypink}0.81+-0.02                    & 0.3+-0.02            
           & 0.74+-0.02                    & 0.26+-0.02                 \\
           & 2000    
           & \cellcolor{babypink}0.97+-0.01                    & \cellcolor{cyan}0.59+-0.02        
           & 0.91+-0.01                    & 0.4+-0.02          
           & 0.92+-0.01                    & \cellcolor{cyan}0.6+-0.02           
           & 0.93+-0.01                    & 0.54+-0.02          
           & 0.92+-0.01                     & 0.49+-0.02         
           & 0.9+-0.02                     & 0.44+-0.02                 \\
           & 5000    
           & \cellcolor{babypink}0.99+-0                       & \cellcolor{cyan}0.68+-0.01        
           & 0.97+-0.01                    & 0.49+-0.02          
           & 0.97+-0.01                    & \cellcolor{cyan}0.67+-0.01            
           & 0.95+-0.01                    & 0.62+-0.02            
           & 0.94+-0.01                    & 0.63+-0.01            
           & 0.96+-0.01                    & 0.6+-0.02                  \\ \hline
80-5-4-1   & 100    
& 0.36+-0.01                    & \cellcolor{cyan}0.24+-0.01           
& 0.35+-0.01                    & 0.2+-0.01             
& NA                    & NA
& \cellcolor{babypink}0.52+-0.02                    & 0.16+-0.01                
& 0.39+-0.01                    & 0.18+-0.01               
& 0.41+-0.02                    & 0.12+-0.01                 \\
           & 500   
           & \cellcolor{babypink}0.83+-0.01            & \cellcolor{cyan}0.44+-0.01
           & 0.74+-0.02                    & 0.3+-0.01           
           & NA                            & NA
           & 0.8+-0.01                     & 0.38+-0.01           
           & \cellcolor{babypink}0.82+-0.02                    & 0.28+-0.01           
           & 0.72+-0.02                    & 0.24+-0.01 \\
           & 2000    
           & \cellcolor{babypink}0.97+-0.01     & \cellcolor{cyan}0.58+-0.01        
           & 0.92+-0.01                    & 0.43+-0.02          
           & NA                    & NA
           & 0.89+-0.01                    & 0.54+-0.01          
           & 0.93+-0.01                     & 0.49+-0.01         
           & 0.93+-0.01                     & 0.46+-0.01                 \\
           & 5000    
           & \cellcolor{babypink}0.99+-0         & \cellcolor{cyan}0.65+-0.01        
           & 0.96+-0.01                    & 0.51+-0.01          
           & NA                    & NA
           & 0.91+-0.01                    & 0.62+-0.01            
           & 0.94+-0.01                    & \cellcolor{cyan}0.64+-0.01            
           & 0.79+-0.02                    & 0.52+-0.02                  \\ \hline           
\end{tabular}
}
\end{center}
\end{table}
\end{landscape}

\begin{figure}[p]
  \centering
    \includegraphics[scale=0.6]{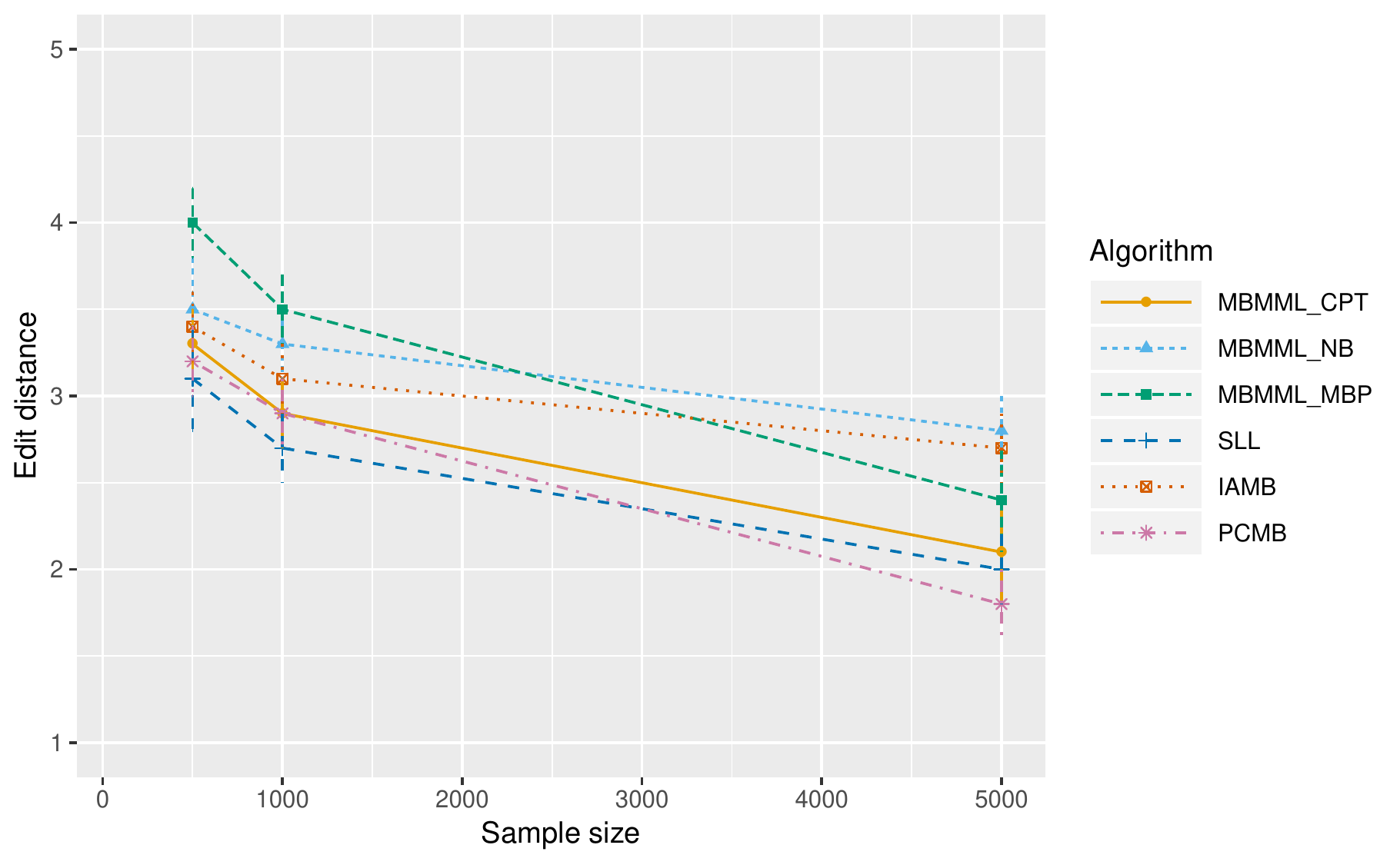}
  \caption{Edit distance (with $95\%$ confidence intervals) vs. sample size on INSURANCE network.}
\end{figure}

\begin{figure}[p]
  \centering
    \includegraphics[scale=0.6]{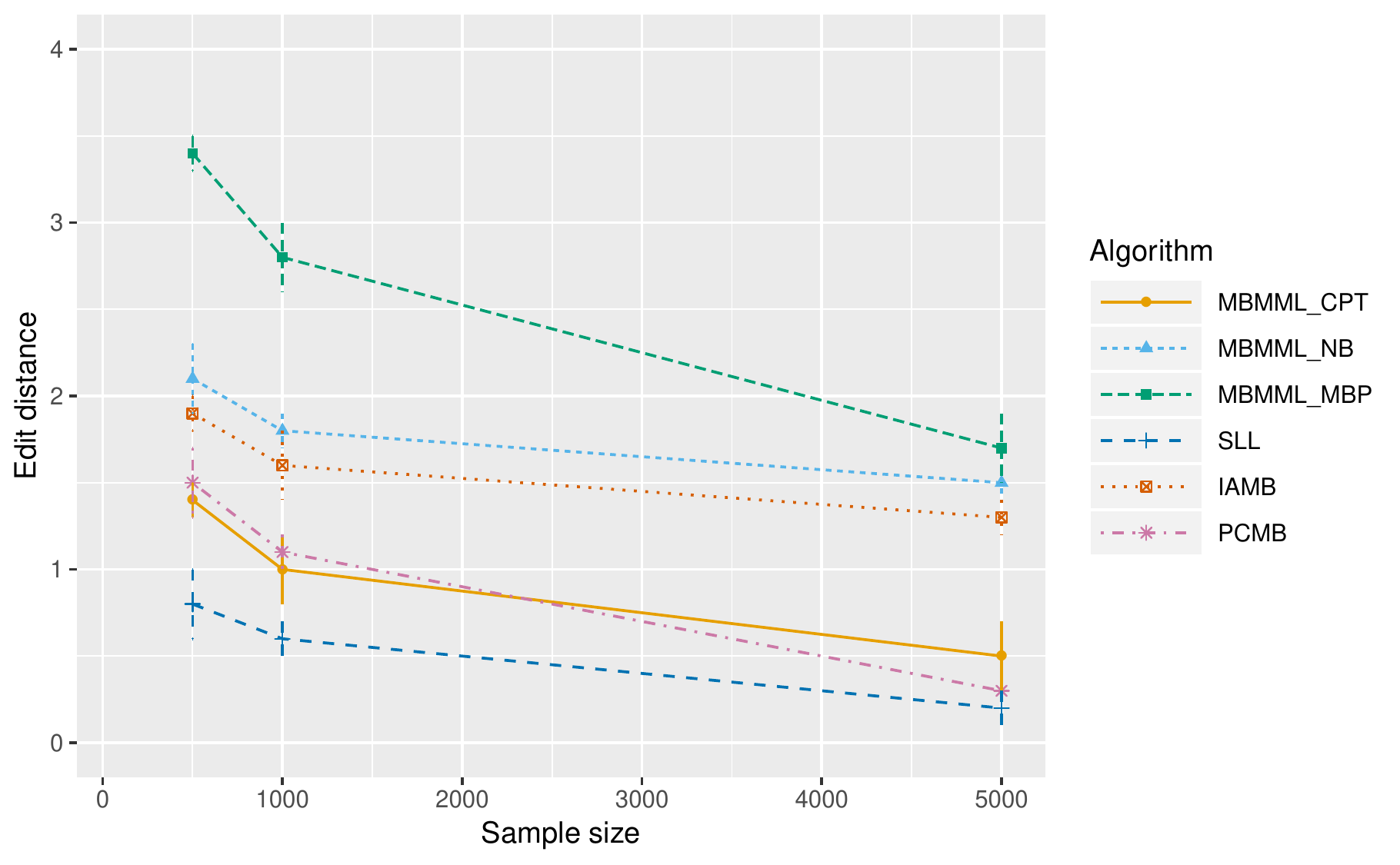}
  \caption{Edit distance (with $95\%$ confidence intervals) vs. sample size on ALARM network.}
\end{figure}

\begin{figure}[p]
  \centering
    \includegraphics[scale=0.6]{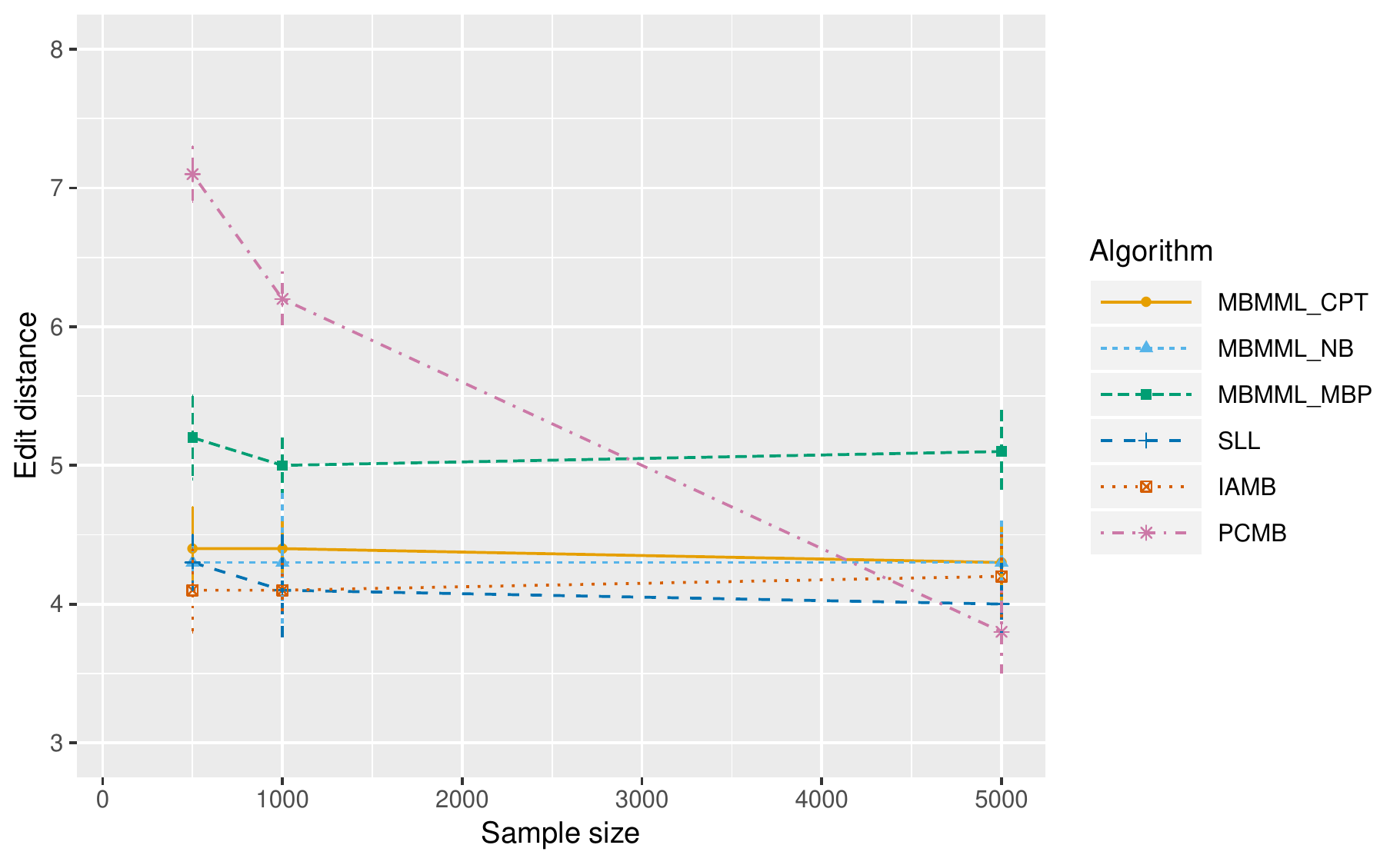}
  \caption{Edit distance (with $95\%$ confidence intervals) vs. sample size on HAILFINDER network.}
\end{figure}

\begin{figure}[p]
  \centering
    \includegraphics[scale=0.6]{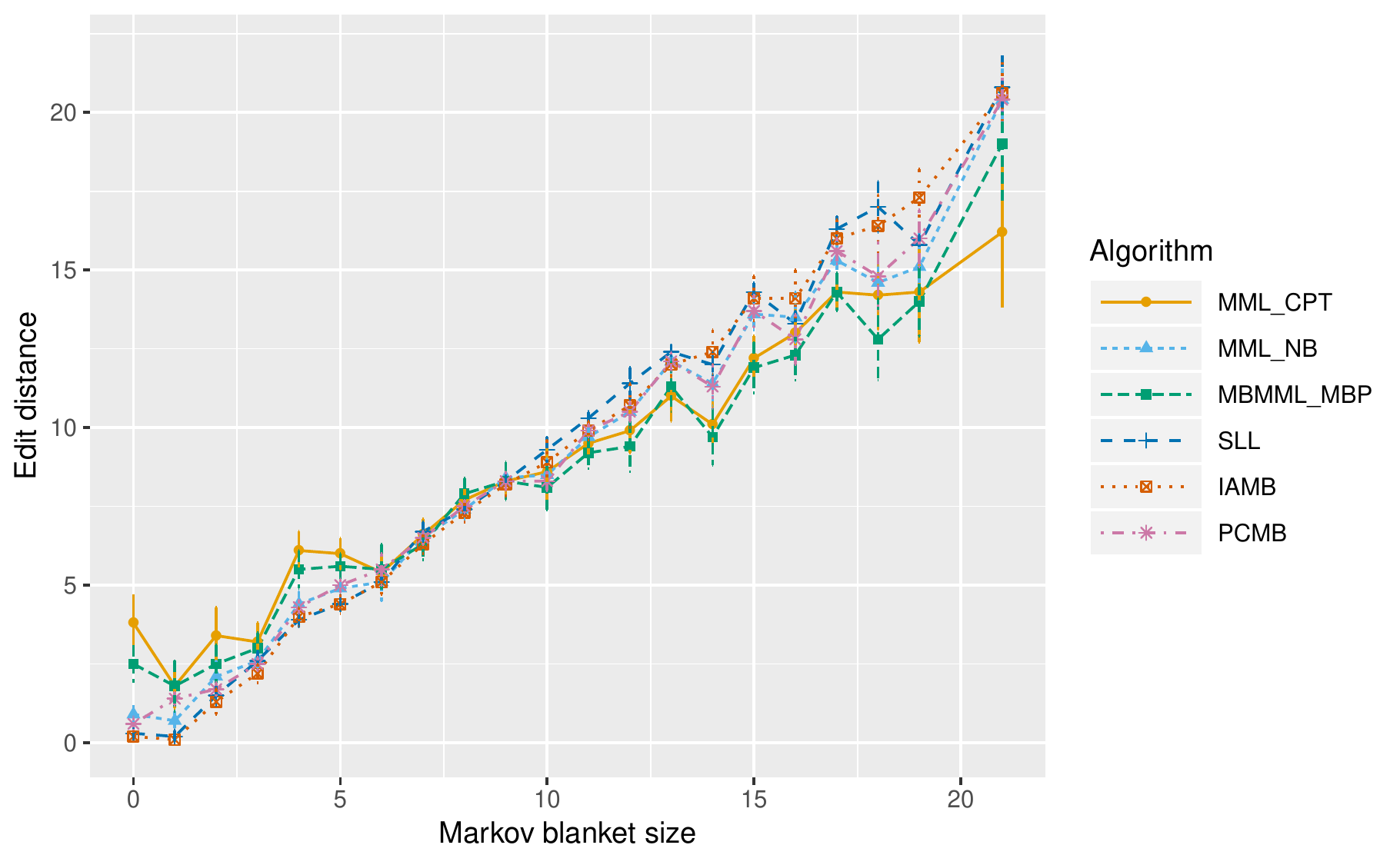}
  \caption{Edit distance against Markov Blanket size on 30-5-4-1 models with 100 samples.}
\end{figure} 

\begin{figure}[p]
  \centering
    \includegraphics[scale=0.6]{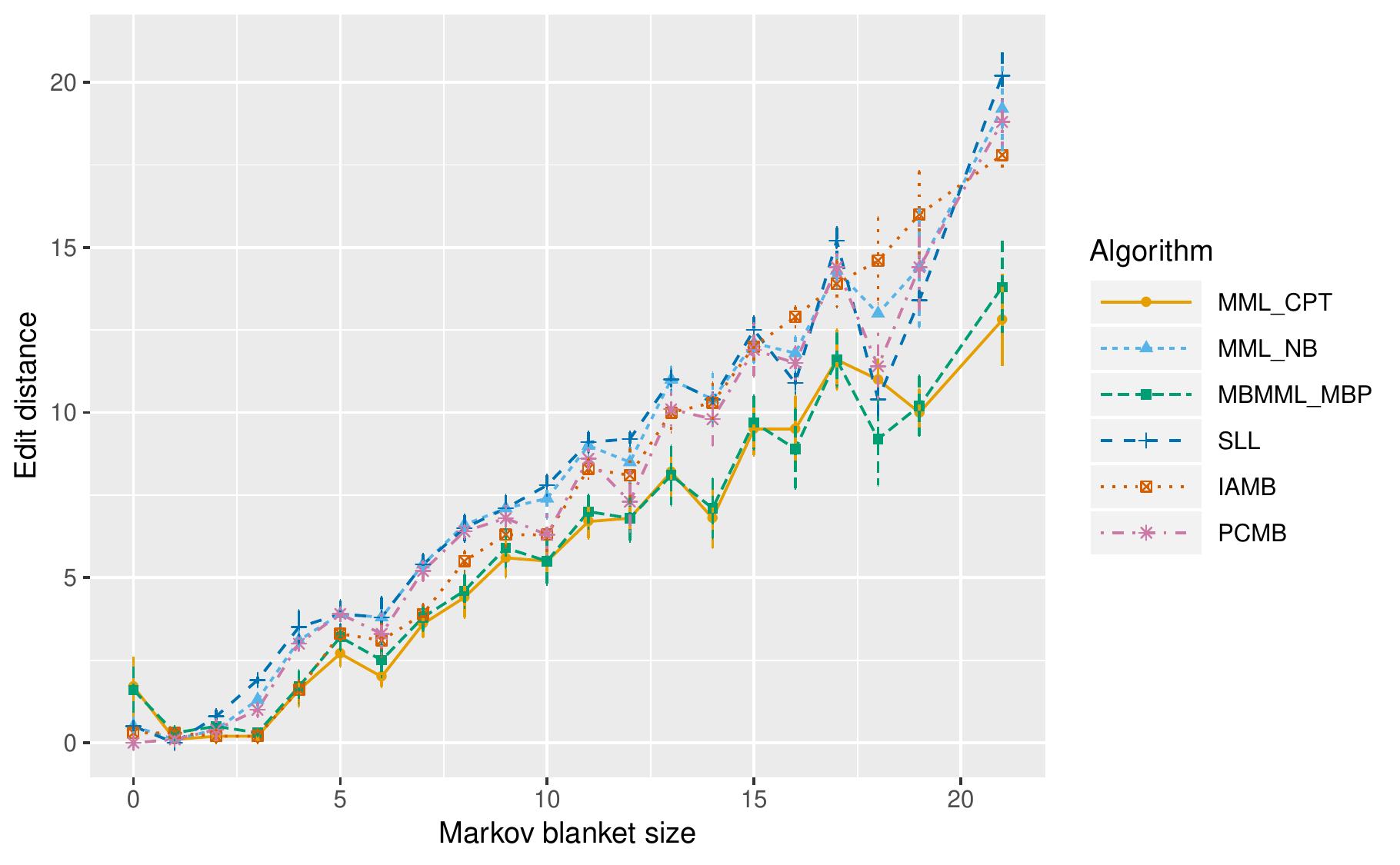}
  \caption{Edit distance against Markov Blanket size on 30-5-4-1 models with 500 samples.}
\end{figure}

\begin{figure}[p]
  \centering
    \includegraphics[scale=0.6]{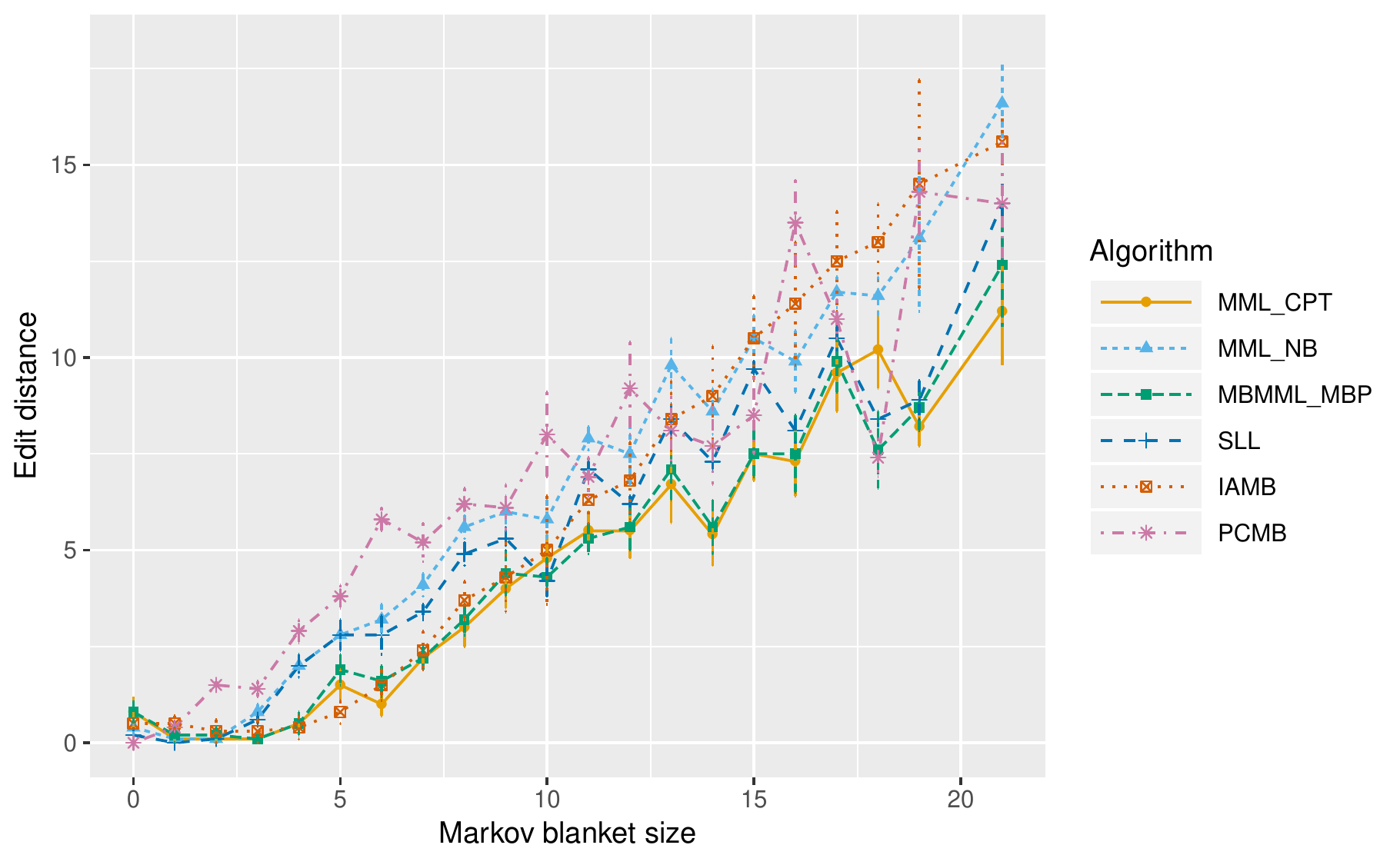}
  \caption{Edit distance against Markov Blanket size on 30-5-4-1 models with 2000 samples.}
\end{figure} 

\begin{figure}[p]
  \centering
    \includegraphics[scale=0.6]{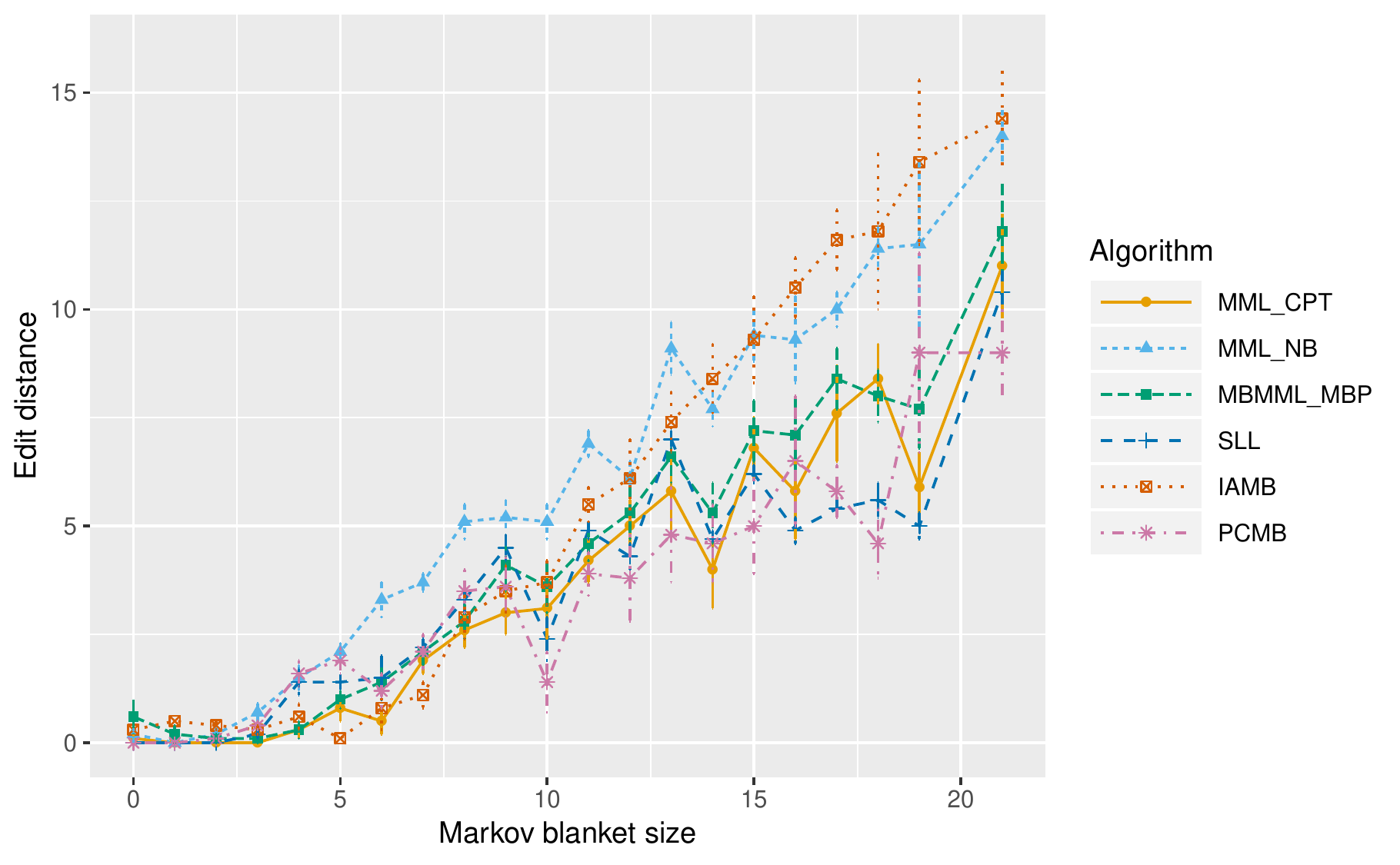}
  \caption{Edit distance against Markov Blanket size on 30-5-4-1 models with 5000 samples.}
\end{figure} 

\begin{figure}[p]
  \centering
    \includegraphics[scale=0.6]{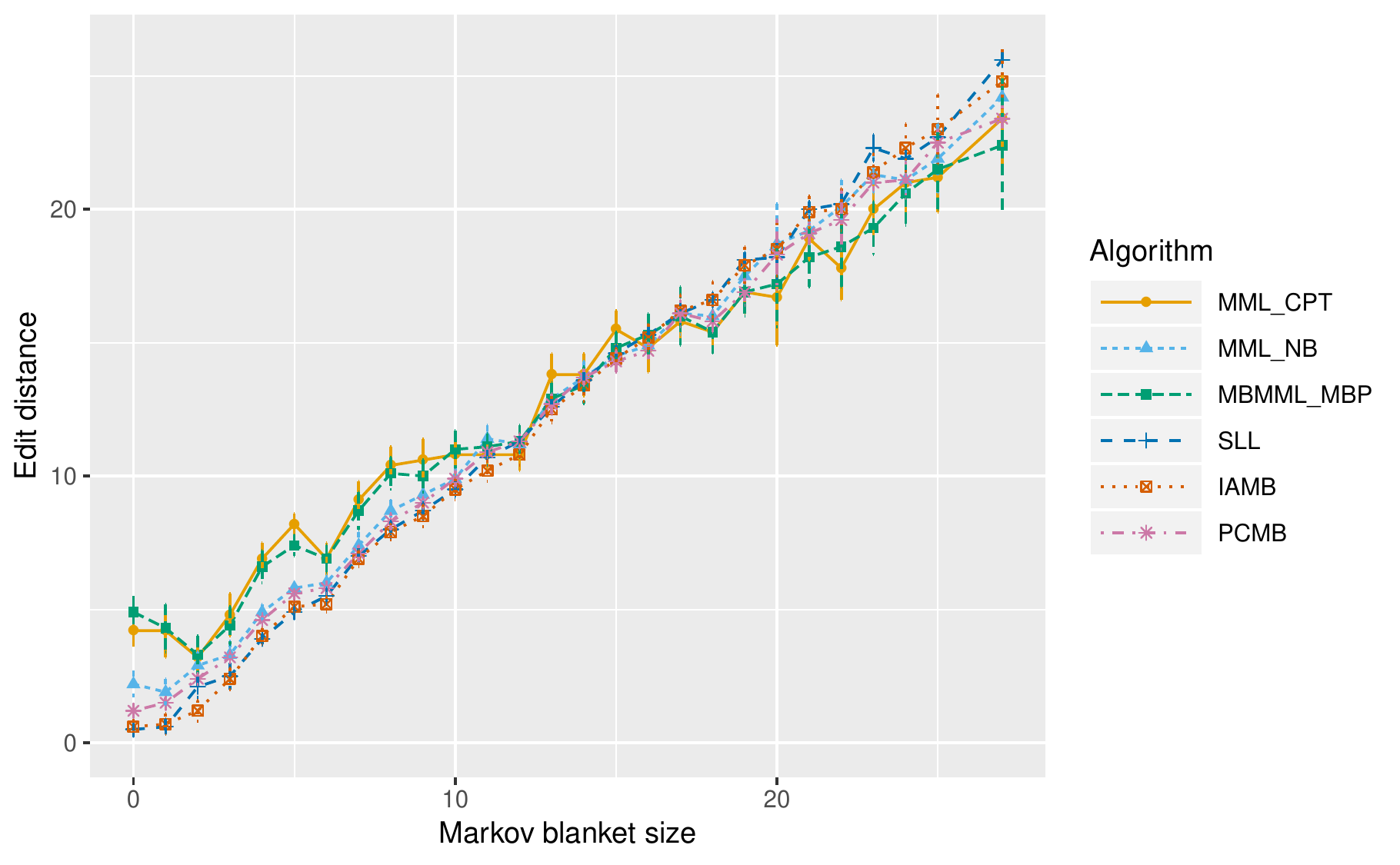}
  \caption{Edit distance against Markov Blanket size on 50-5-4-1 models with 100 samples.}
\end{figure} 

\begin{figure}[p]
  \centering
    \includegraphics[scale=0.6]{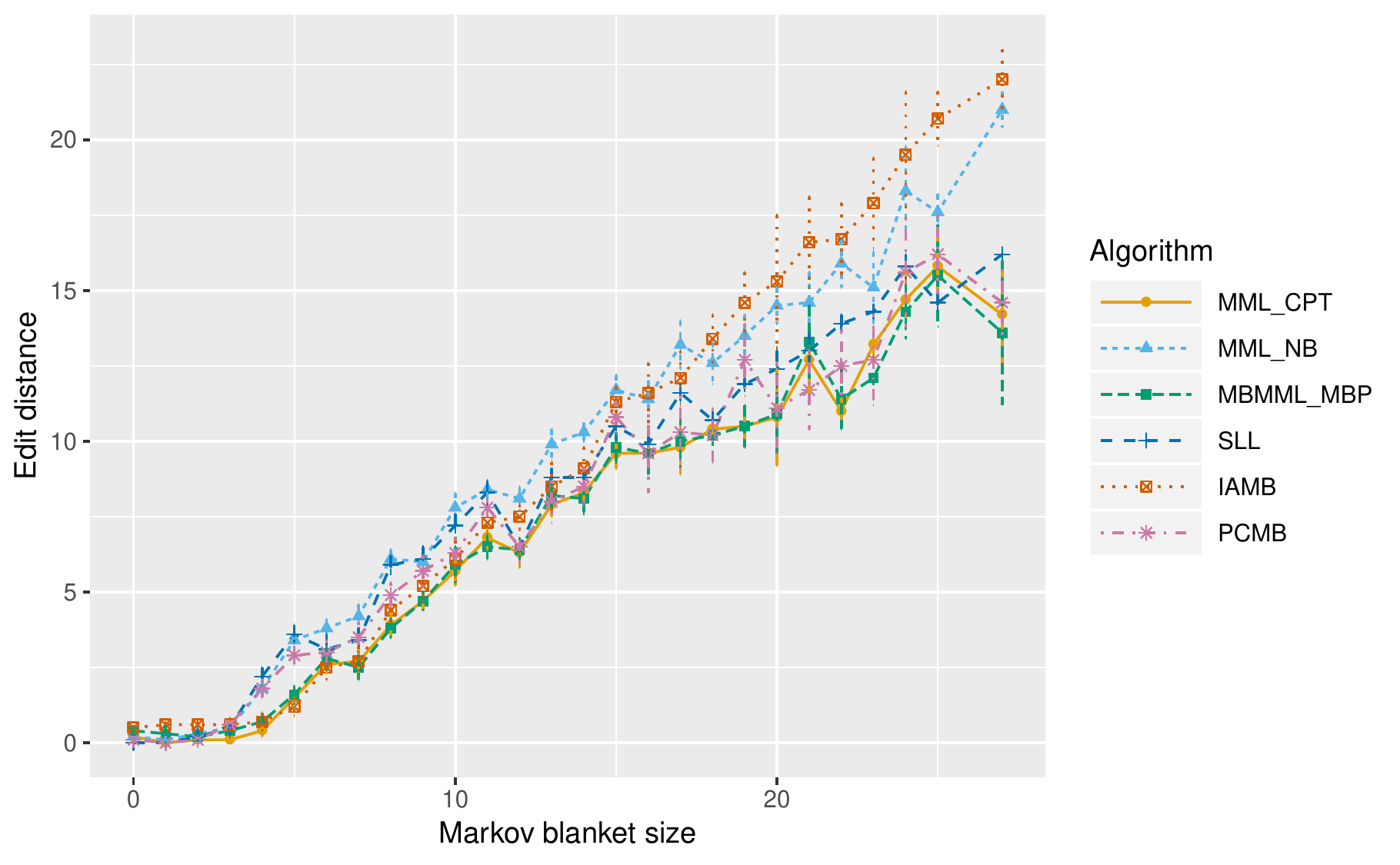}
  \caption{Edit distance against Markov Blanket size on 50-5-4-1 models with 2000 samples.}
\end{figure}

\newpage
\bibliography{mb_discovery}

\end{document}